\documentclass[12pt]{article}
\DeclareUnicodeCharacter{0301}{\'{e}}
\usepackage{amssymb}
\usepackage{amsmath}
\usepackage{amsthm}
\usepackage{color}
\usepackage{comment}
\usepackage{enumitem}		
\usepackage{geometry}		
\usepackage{graphicx}
\usepackage{authblk}
\usepackage{amssymb}
\usepackage{listings}
\usepackage{xcolor}
\usepackage{caption}
\usepackage{color}
\usepackage{algorithm}
\usepackage{algpseudocode}
\usepackage{booktabs}
\usepackage{subcaption}

\makeatletter
	\@addtoreset{equation}{section}
\makeatother

\let\originalleft\left
\let\originalright\right
\renewcommand{\left}{\mathopen{}\mathclose\bgroup\originalleft}
\renewcommand{\right}{\aftergroup\egroup\originalright}

\newlist{romanlist}{enumerate}{3}
\setlist[romanlist]{label=\roman*),ref=(\roman*)}

\begin{document}

\newcommand{\cF}{\mathcal{F}}
\newcommand{\cP}{\mathcal{P}}
\newcommand{\cR}{\mathcal{R}}
\newcommand{\cS}{\mathcal{S}}
\newcommand{\cT}{\mathcal{T}}
\newcommand{\ee}{\varepsilon}
\newcommand{\rD}{{\rm D}}
\newcommand{\re}{{\rm e}}

\newtheorem{theorem}{Theorem}[section]
\newtheorem{corollary}[theorem]{Corollary}
\newtheorem{lemma}[theorem]{Lemma}
\newtheorem{proposition}[theorem]{Proposition}

\theoremstyle{definition}
\newtheorem{definition}{Definition}[section]


\title{
Deep Learning for Prediction and Classifying the Dynamical Behaviour of Piecewise Smooth Maps
}
\author[1]{Vismaya V S}
\author[1]{Bharath V Nair}
\author[1]{Sishu Shankar Muni}
\affil[1]{School of Digital Sciences,\\ Digital University Kerala\\
Thiruvananthapuram, PIN 695317, Kerala, India}

\maketitle


\begin{abstract}
This paper explores the prediction of the dynamics of piecewise smooth maps using various deep learning models. We have shown various novel ways of predicting the dynamics of piecewise smooth maps using deep learning models. Moreover, we have used machine learning models such as Decision Tree Classifier, Logistic Regression, K-Nearest Neighbor, Random Forest, and Support Vector Machine for predicting the border collision bifurcation in the \(1D\) normal form map and the \(1D\) tent map. Further, we classified the regular and chaotic behaviour of the \(1D\) tent map and the \(2D\) Lozi map using deep learning models like Convolutional Neural Network (CNN), ResNet50, and ConvLSTM via cobweb diagram and phase portraits. We also classified the chaotic and hyperchaotic behaviour of the \(3D\) piecewise smooth map using deep learning models such as the Feed Forward Neural Network (FNN), Long Short-Term Memory (LSTM), and Recurrent Neural Network (RNN). Finally, deep learning models such as Long Short-Term Memory (LSTM) and Recurrent Neural Network (RNN) are used for reconstructing the two parametric charts of \(2D\) border collision bifurcation normal form map. 
\end{abstract}

\section{Introduction}

In the study of complex systems, it is important to understand how the state of that system changes with time to predict and explain the dynamical behaviour of the system. A dynamical system is a system that changes and \cite{Rencontres1975} evolves over time. For a wide range of applications ranging from predicting weather patterns, controlling complex machinery, and modelling biological systems \cite{Lathrop2015}, understanding the behaviour of dynamical systems is important. In addition, the analysis of dynamical systems has applications in fields such as economics \cite{HSIEH1991}, finance \cite{Cheriyan2016}, ecology \cite{Bieg2017}, and engineering \cite{Stefano}, where the study of system behaviour can lead to the development of more efficient and effective strategies. For instance, in the field of ecology, the study of population dynamics has been crucial in predicting and managing the impact of human activities on wildlife populations. Dynamical systems theory \cite{Rencontres1975} is a field that provides the mathematical tools to analyse the evolution of such systems, which can exhibit either continuous or discrete behaviour.

In many domains, such as dynamical systems theory \cite{diBernardo2001}, physics \cite{Banerjee2000}, and engineering, predicting and classifying various dynamical behaviour of piecewise smooth maps is important. Piecewise smooth (PWS) maps \cite{mario} are mathematical models showing continuous and discontinuous behaviours. At the boundaries between the regions, the map shows discontinuities and continuous behaviour within each region. The combination of these continuous regions and discontinuities can lead to complex dynamics, including phenomena like bifurcations \cite{DiBernardo2002}, chaos \cite{Patra2018}, etc. There are different types of PWS systems. They are classified based on Degree of Smoothness(DoS). When DoS equal to 1, such systems are called impacting systems or piecewise smooth hybrid systems \cite{Stolyarov2022}, for example, oscillators in machines. When DoS equal 1, such systems are called Filippov systems \cite{Kuznetsov2003}, for example, relay feedback systems. When DoS equal to 2 or more, such systems are called piecewise smooth continuous systems.
Border collision bifurcation \cite{Banerjee1999} is a distinct type of bifurcation occurring in piecewise smooth maps when a fixed point, periodic point, chaotic attractor, or any attractor collides with the boundary region. Border collision bifurcation \cite{Nusse1994} occurs when the parameter of the system is varied, resulting in a fixed point or periodic orbit colliding with the boundary, and a discontinuous change will occur. Most of the bifurcations seen in the power converters are border collision bifurcations \cite{GuohuiYuan1998}. Unlike other bifurcations like saddle-node (fold) bifurcation \cite{Inoue2014}, transcritical bifurcation \cite{Ferrer2022}, pitchfork bifurcation \cite{Rajapakse2017}, etc, border collision bifurcation can lead to sudden changes. For instance, a stable fixed point can suddenly become unstable and bifurcate to a chaotic attractor \cite{Didov2020}. This feature is not seen in smooth maps. Border collision bifurcations are important to understand the evolution of piecewise smooth systems \cite{Nusse1992} \cite{NUSSE1995}, especially in the context of stability analysis and control systems. For example, a study in the dynamics of automatic control systems with pulse-width modulation of the second kind (PWM-2) \cite{Zhusubaliyev}. Predicting border collision bifurcations is important for controlling the dynamical behaviour of the systems that exhibit piecewise smooth dynamics \cite{ZHUSUBALIYEV2001}. In recent decades, studies have been conducted to understand the bifurcations in piecewise smooth maps \cite{Robert2002}. 

The piecewise smooth map can show a wide range of dynamical behaviours such as regular \cite{Hogan2006}, chaotic \cite{Avrutin2014} and hyperchaotic behaviours \cite{Patra2020}. Understanding the behaviour like regular, chaotic \cite{ZHUSUBALIYEV2001} and hyperchaos are important for various applications like control systems \cite{Ahmad}, cryptography \cite{Pareek2005}, etc. In engineering systems like robotic arms,  mechanical systems with impacts, and electrical circuits with switching elements are modelled using piecewise smooth maps \cite{mario}. In control systems, piecewise smooth maps are used to model and analyze systems with switching controllers such as sliding mode control \cite{Sundarapandian}. Piecewise smooth models are also used in signal processing to handle signals with abrupt changes and discontinuity \cite{Storath2023}. They are also used in biological systems like understanding heart rhythms \cite{Rosenblum1995}, brain functions \cite{COOMBES2018}, in economics and social sciences \cite{Tramontana2012} and in environmental models such as climate predictions, etc.

Using machine learning to predict the border collision bifurcation is a novel approach \cite{Bury2023}. Nowadays, Deep Learning techniques are becoming increasingly essential in various fields, including climate forecasting \cite{Salman2015}, healthcare \cite{Liang2014}, virtual assistants \cite{Someshwar2020}, finance \cite{Sezer2020}, e-commerce \cite{Shankar}, agriculture \cite{Nguyen2020}, automotive \cite{Singh} and transportation \cite{GuerreroIbaez2020}, security \cite{Aversano2021}, etc. Deep learning has emerged as a powerful tool in the field of dynamical systems \cite{Rajendra2020} \cite{LeCun2015}. Traditional methods in dynamical systems frequently depend on analytical methods that might be difficult to handle complex, high-dimensional, or nonlinear processes. Deep learning techniques like Feed Forward Neural Network (FNN) \cite{Bebis1994}, Long short-term Memory (LSTM) \cite{Hochreiter1997}, Recurrent Neural Network (RNN) \cite{Du2013}, and Convolutional Neural Network (CNN) \cite{Ketkar2021} are some of the techniques commonly used. 
Deep Learning techniques are useful in predicting and forecasting the future state of dynamcial systems like weather forecasting \cite{Hewage2020}. However, analysing vast datasets, especially those involving time series data, presents notable challenges.

The present paper explores predicting the dynamics of piecewise smooth maps like border collision bifurcation using machine learning models such as Decision Tree Classifier, Logistic Regression,
K-Nearest Neighbor, Random Forest and Support Vector Machine and classification of regular and chaotic behaviour using deep learning models like Convolutional Neural Network (CNN), ResNet50, and ConvLSTM via cobweb diagram and phase portraits. For the classification of chaotic and hyperchaotic behaviour using deep learning techniques, various techniques are used, such as Feed Forward Neural Network (FNN), Recurrent Neural Network (RNN), Long short-term Memory (LSTM), Convolutional Neural Network (CNN), ResNet50, etc. Finally, we used deep learning models to reconstruct the two parameter charts. The study shows that it is important to understand the dynamics of the system for application in various fields.

The novelty of the paper includes the following:
\begin{itemize}

    \item Regular, chaotic and hyperchaotic dynamical behaviour of piecewise smooth maps are classified using deep learning models like FeedForward Neural Network (FNN), Long Short-Term Memory (LSTM) and Recurrent Neural Network (RNN).
    
      \item Beyond deep learning models, the paper also       
      includes machine learning models such as Decision Tree Classifier, Logistic Regression, K-Nearest Neighbor, Random Forest and Support Vector Machine for the border collision bifurcation prediction tasks.
     \item In the paper, machine learning models are used for predicting the higher periods of the periodic orbits.
     \item Deep learning models employ cobweb and phase portraits of \(1D\) tent map and \(2D\) Lozi map to classify regular and chaotic dynamical behaviour.
     \item Deep learning models are used to reconstruct the two-parameter charts in the paper.
\end{itemize}

The paper is organised as follows: Section \ref{section2} explores piecewise smooth maps like \(1D\) border collision normal form map and the \(1D\) tent map and predicts the border collision bifurcations. Section \ref{section3} includes the prediction of border collision bifurcation using different machine learning models, which include Decision Tree Classifier, Logistic Regression, K-Nearest Neighbor, Random Forest and Support Vector Machines. In section \ref{section4}, the classification of the chaotic and regular behaviour of the \(1D\) tent map and the \(2D\) lozi map is done using deep learning models, namely Convolutional Neural Network (CNN), ResNet50, and ConvLSTM. In Section \ref{section5}, the classification of the chaotic and hyperchaotic behaviour of \(3D\) piecewise smooth map using deep learning models like FeedForward Neural Networks (FNN), Long Short-Term Memory (LSTM) networks, and Recurrent Neural Networks (RNN), is discussed. In section \ref{section6}, two parameter charts are reconstructed using deep learning models. Finally, Section \ref{section7} includes the conclusion and future scopes.

\section{Border Collision Bifurcation in Piecewise Smooth Map} \label{section2}
Here, we considered the two simplest piecewise smooth maps, i.e., \(1D\) normal form map and \(1D\) tent map.
\subsection{Normal Form of One-Dimensional Piecewise Smooth Map} \label{section1.1}
The normal form of \(1D\) piecewise smooth map \cite{JAIN2003} is defined as follows:
\begin{equation}
x_{n+1}=
    \begin{cases}
        ax_n+\mu  & \text{for } x_n < 0\\
        bx_n+\mu+l & \text{for } x_n > 0,
    \end{cases}
    \label{normaleq}
\end{equation}
where \(x_n\) represents the state of the system at time step \(n\), \(x_{n+1}\) represents the state of the system at time step \(n+1\), \(\mu\) represents the parameter affecting the system's behavior, and \(l\) represents the length of discontinuity.

\(L\) (left) and \(R\) (right) are the two sides of the state space in the normal form map. The fixed point \(L\) is located at 
\begin{equation}
\begin{aligned}
x_L^* = \frac{\mu}{1-a}
\end{aligned}
\end{equation}
and the fixed point \(R\) is located at
\begin{equation}
\begin{aligned} 
x_R^* &= \frac{\mu+l}{1-b}
\end{aligned}
\end{equation}
When \(\mu=0\), the fixed point \(x_L^*\) collides with the border at \(x=0\). And when \(\mu=l\), the fixed point \(x_R^*\) collide with the border at \(x=0\). As a result, when the parameter \(\mu\) is changed, we should anticipate two border collision events at parameter \(\mu=0\) and \(\mu=l\).
\begin{figure}[!hbtp]
    \centering
    \includegraphics[width=0.8\linewidth]{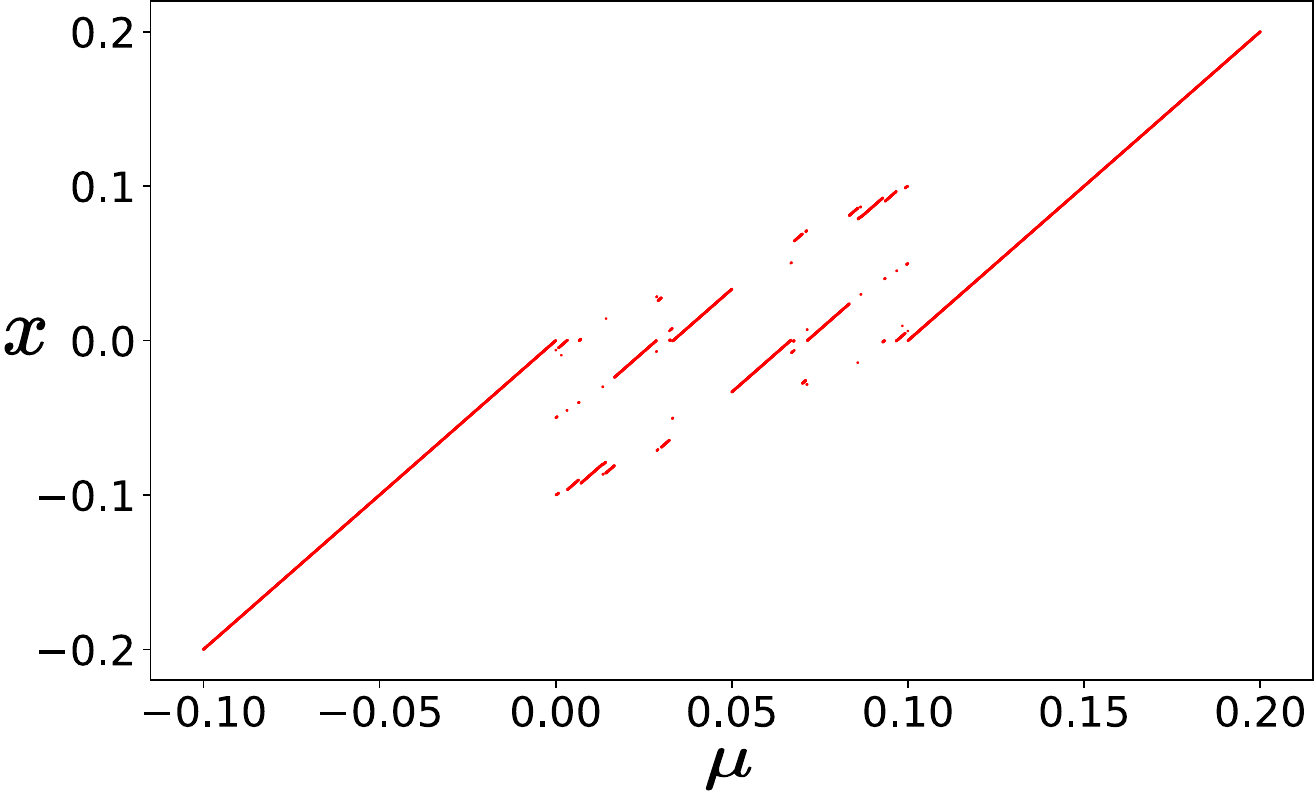}
    \caption{One-parameter border collision bifurcation diagram of normal form map \eqref{normaleq} where \(x\)-axis presents the parameter \(\mu\) and \(y\)-axis represents the state variable \(x\), \(\mu=(-0.1,2)\).}
    \label{bbnormal}
\end{figure}

Fig:\ref{bbnormal} shows the one-parameter border collision bifurcation diagram of the normal form map \eqref{normaleq} at \(a=0.5\),\(b=0.5\), \(l=-0.1\) and \(\mu\) $\in$ \((-0.1,0.2)\) where \(x\)-axis represents the parameter \(\mu\) and \(y\)-axis represents the state variable \(x\). The figure shows that the behaviour of the normal form map changes as the parameter \(\mu\) is varied from \(-0.1\) to \(0.2\). The location of the fixed point varies as the parameter \(\mu\) is changed between \(-0.1\) and \(0.2\). At specific \(\mu\) values, new dynamical behaviours occur as a result of the fixed points colliding with the border, i.e., \((x=0)\). For example, when \(\mu=-0.1\), there is a single stable fixed point as in fig:\ref{bbnormal}. As \(\mu\) increased, the fixed point changes linearly with \(\mu\) and becomes unstable at \(\mu=0\). A sequence of dots on the diagram shows the emergence of a new periodic orbit at the border collision bifurcation point. The periodic orbit experiences several bifurcations like period-doubling bifurcation as \(\mu\) increases, resulting in additional periodic orbits and chaotic behaviour. Based on the value of the parameter \(\mu\), the diagram shows the position and stability of various orbits.

Because of the presence of the periodic behaviour, we want to know the period of the orbit \(x\) for a particular value of the parameter \(\mu\). It's important to know if there is a periodic behaviour. Detecting the periodic behaviour includes simulating the system's behaviour and then observing the trajectory. If the trajectory returns close to its starting point after a certain number of iterations, then it shows a periodic behaviour. To determine the period, we will count the number of iterations required to complete one cycle. In this way, we can determine the period of the orbit \(x\) for a particular parameter value \(\mu\).

\begin{figure}[!hbtp]
    \centering
    \includegraphics[width=0.8\linewidth]{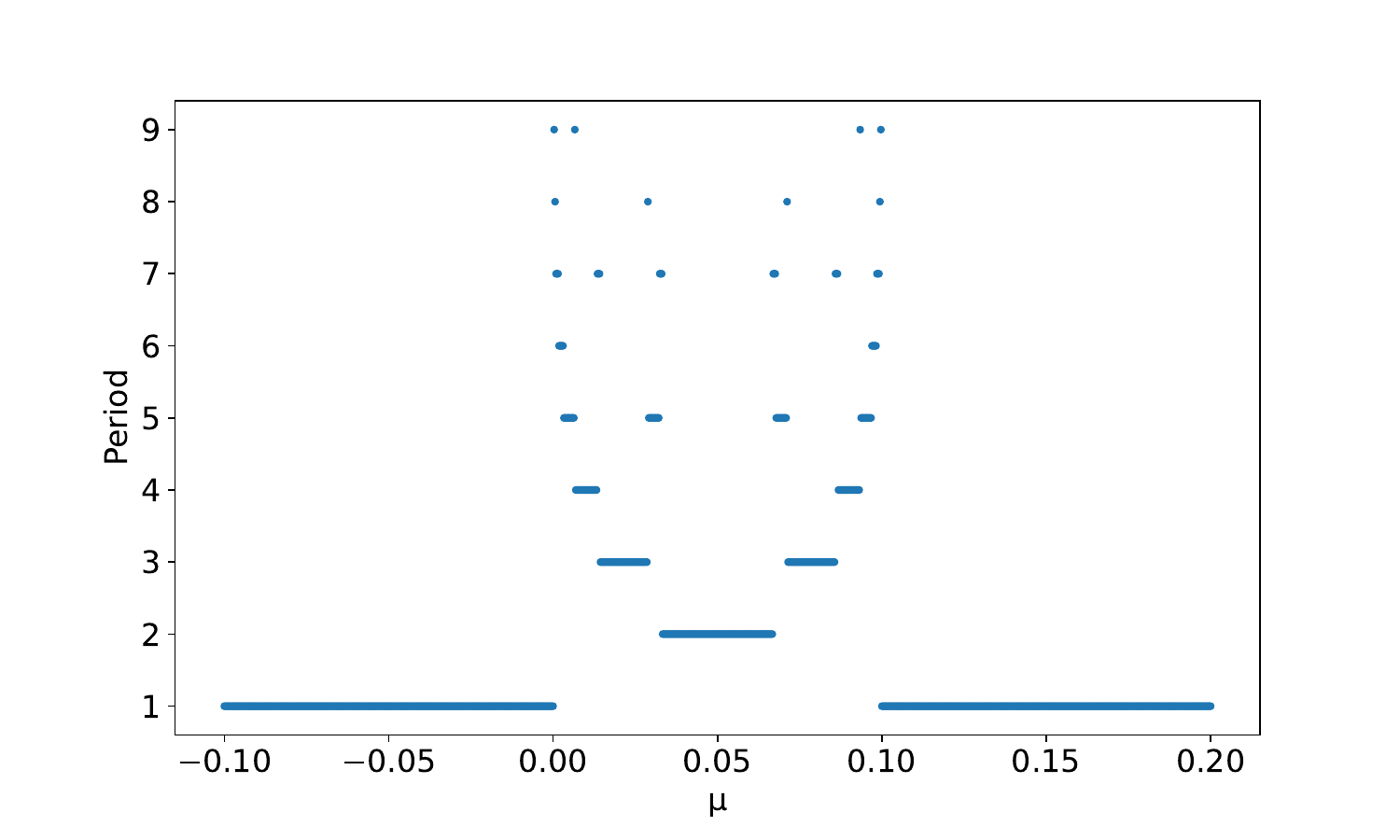}
    \caption{One-parameter Period vs \(\mu\) diagram of normal form map \eqref{normaleq} where \(x\)-axis represents \(\mu\) values and \(y\)-axis represents periods, \(\mu\) varies from \(-0.10\) to \(0.20\) and period varies from \(1\) to \(9\).}
    \label{pvsmunormal}
\end{figure}
Figure \ref{pvsmunormal} shows a one-parameter diagram of period vs \(\mu\). The \(x\)-axis represents the value of the parameter \(\mu\), and the \(y\)-axis represents the period. In the figure, the period of the orbit \(x\) for every value of \(\mu\) is represented by blue. The sudden changes in the period at specific values of \(\mu\) lead to a sequence of discrete points, which is why the line is not smooth. The figure displays multiple areas with varying dynamical behaviours as the parameter \(\mu\) changes. For \(\mu\) starting from \(-0.1\), the period is constant, i.e., \(1\), but when the value of parameter \(\mu\) increases, the period suddenly increases to \(9\), this happens because of the occurrence of the border collision bifurcation. Then, the period varies till it approaches \(\mu=0.1\), then it suddenly decreases to \(\mu=1\) because of the presence of several border collision bifurcations. 

\begin{figure}[!hbtp]
    \centering
    \includegraphics[width=0.8\linewidth]{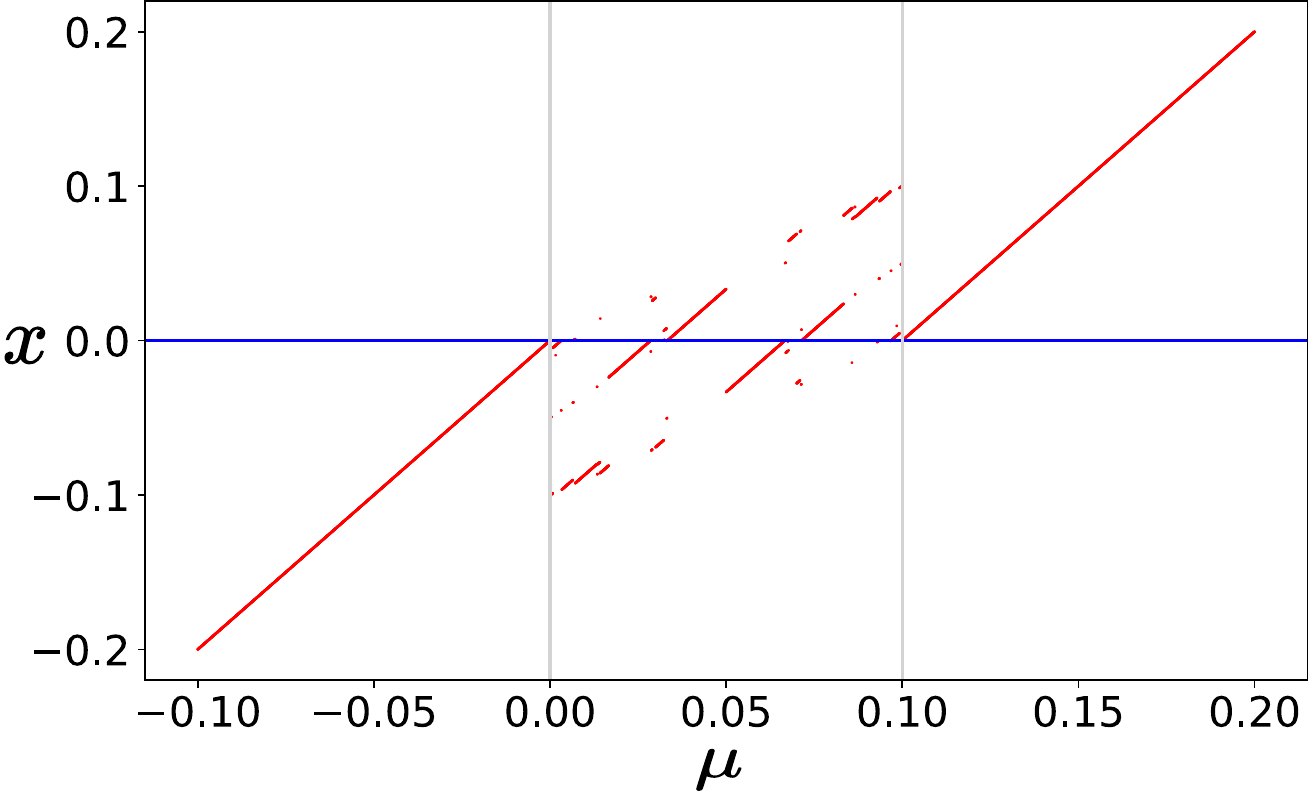}
    \caption{One-parameter border collision bifurcation diagram of normal form map \eqref{normaleq} showing the \(\mu\) values at which border collision bifurcation occur i.e., \(\mu=1.3877787807814457e-17\) and \(\mu=0.10000000000000003\) denoted by grey vertical lines.}
    \label{bcbdiagramnormalmap}
\end{figure}

Next, we want to know at which value of \(\mu\) border collision bifurcation occurs. For convenience, we start with period \(=1\) or fixed point and then apply the condition, i.e., for which the value of \(\mu\), \(x\) converges to \(0\) or collides with the border. Finally, we got two values of \(\mu\) having period \(1\), which shows border collision bifurcation; they are \(\mu=1.3877787807814457e-17\) and \(\mu=0.10000000000000003\) as shown in the Fig:\ref{bcbdiagramnormalmap} denoted by grey vertical lines.
We also observe the occurrence of the border collision bifurcation of periodic orbits of higher periods see fig:\ref{bbnormal}. However, computational techniques are required to predict the border collision bifurcation. Here, we use machine learning techniques to predict the border collision bifurcation of the period orbits having higher periods in Section \ref{section3}.

Next, we try to understand the border collision bifurcation in another prototypical piecewise smooth map, i.e., the \(1D\) tent map.

\subsection{\(1D\) Tent Map}
The \(1D\) tent map is a simple but attractive illustration of a piecewise chaotic dynamical system \cite{Yoshida1983}. The tent map is defined by the following equation:
\begin{equation}
x_{n+1}=
    \begin{cases}
        \mu x_n & \text{for } x_n < 0.5\\
        \mu(1-x_n) & \text{for } x_n > 0.5,
    \end{cases}
    \label{tenteq}
\end{equation}

where \(x_{n+1}\) represents the value of the state variable \(x\) at the next iteration \(n+1\), \(x_n\) represents the current value of the state variable \(x\) at iteration \(n\), and \(\mu\) is the parameter that determines the dynamical behaviour of the map.

\begin{figure}[!hbtp]
    \centering
    \includegraphics[width=0.8\linewidth]{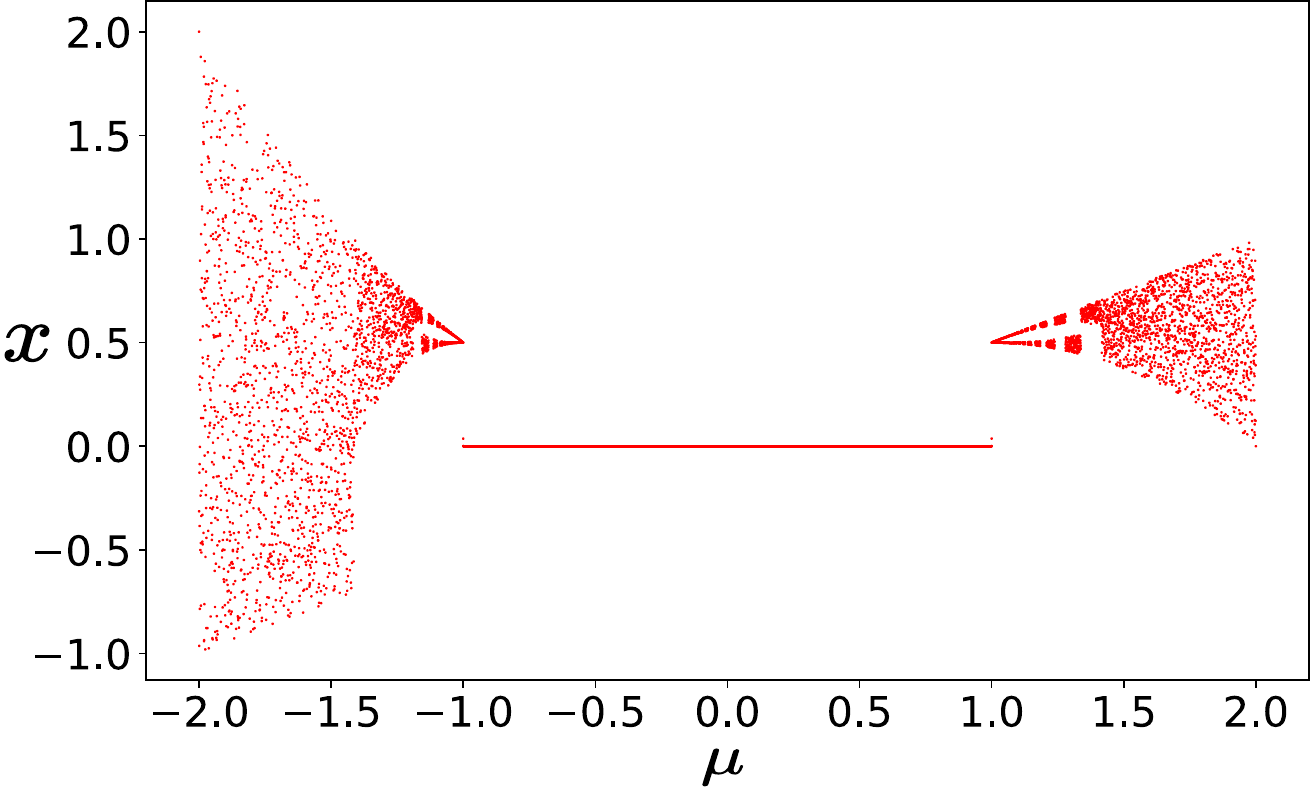}
    \caption{One-parameter border collision bifurcation diagram of the \(1D\) tent map where \(x\)-axis represent the parameter \(\mu\) and \(y\)-axis represent the state variable \(x\), \(\mu=(-1.5,1.5)\).}
    \label{bcbdiagramtent}
\end{figure}

Figure \ref{bcbdiagramtent} shows the one-parameter border collision bifurcation diagram \(\mu\) vs \(x\) of the \(1D\) tent map. The values of the parameter \(\mu\) are represented on the \(x\)-axis and range from \(-1.5\) to \(1.5\). The values of \(x\) are represented on the \(y\)-axis. Starting from an initial value of \(x_0=0.1\), each point on the figure indicates the value of \(x\) after \(10000\) iterations of the tent map. The tent map exhibits different dynamical behaviour when the \(\mu\) value varies from \(-1.5\) to \(1.5\). When \(\mu\) changes from \(-1.5\) to \(-1.0\), the dynamical behaviour of the \(1D\) tent map changes from chaotic to periodic or fixed point and when \(\mu\) varies from \(1.0\) to \(1.5\), the dynamical behaviour again becomes chaotic. 

Because of the presence of periodic behaviour, we want to know the period of orbit \(x\) for a particular value of \(\mu\). Detecting periodic behaviour is similar to the previous technique mentioned in section \ref{section1.1}. In this way, we can determine the period of \(x\) for a particular value of \(\mu\).

\begin{figure}[!hbtp]
    \centering
    \includegraphics[width=0.8\linewidth]{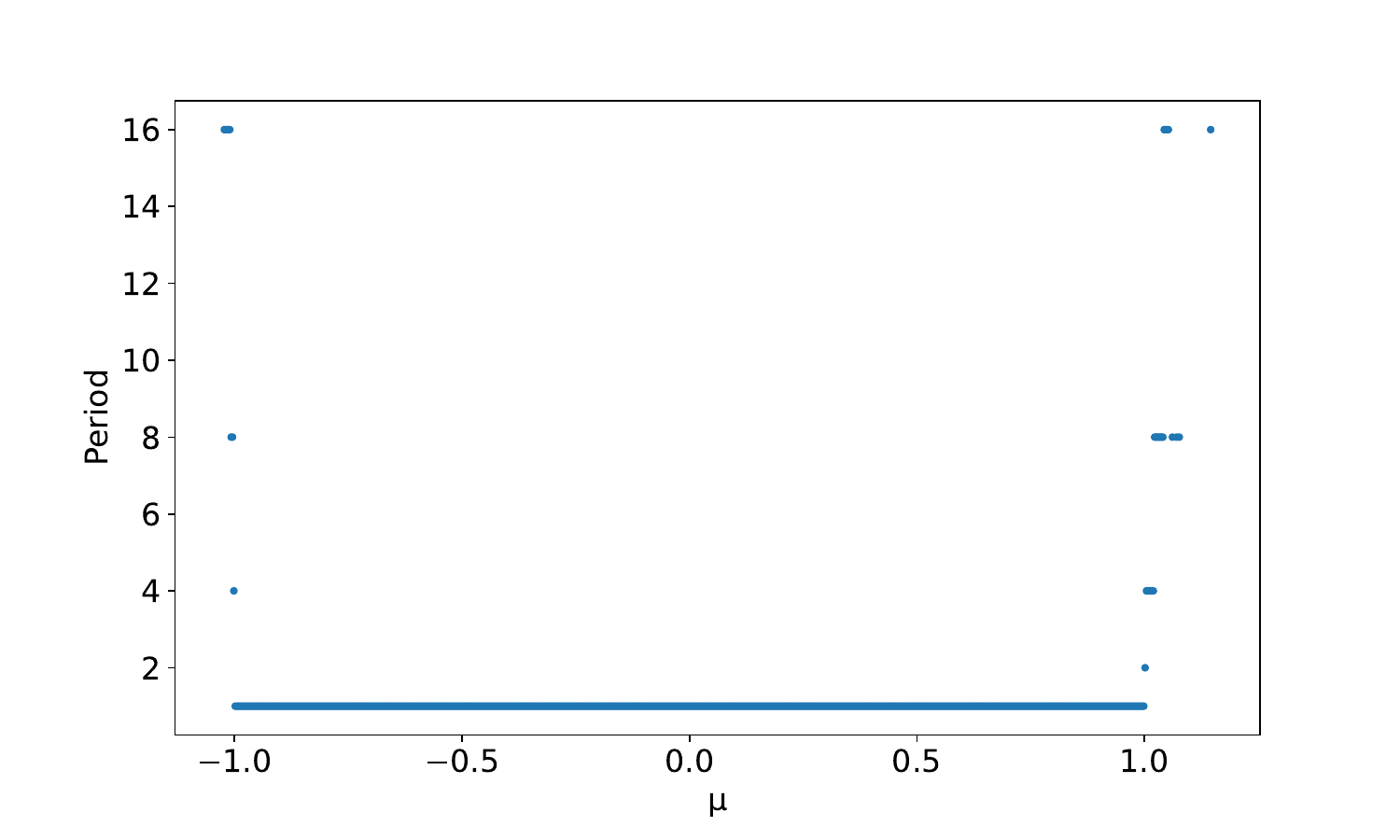}
    \caption{One parameter period vs \(\mu\) diagram of tent map where where \(x\)-axis represents the parameter \(\mu\) values and \(y\)-axis represents the periods, \(\mu\) varies from \(-1.0\) to \(1.0\) and period varies from \(2\) to \(16\).}
    \label{pvsmutent}
\end{figure}

Figure \ref{pvsmutent} shows a one-parameter period vs \(\mu\) diagram that shows the relationship between the parameter \(\mu\) and the period of the \(1D\) tent map. The figure shows that as \(\mu\) varies from \(-1.0\) to \(1.0\), the period increases gradually for each step. There is a sudden increase in the period at certain values of parameter \(\mu\) because of the presence of border collision bifurcation. 

\begin{figure}[!hbtp]
    \centering
    \includegraphics[width=0.8\linewidth]{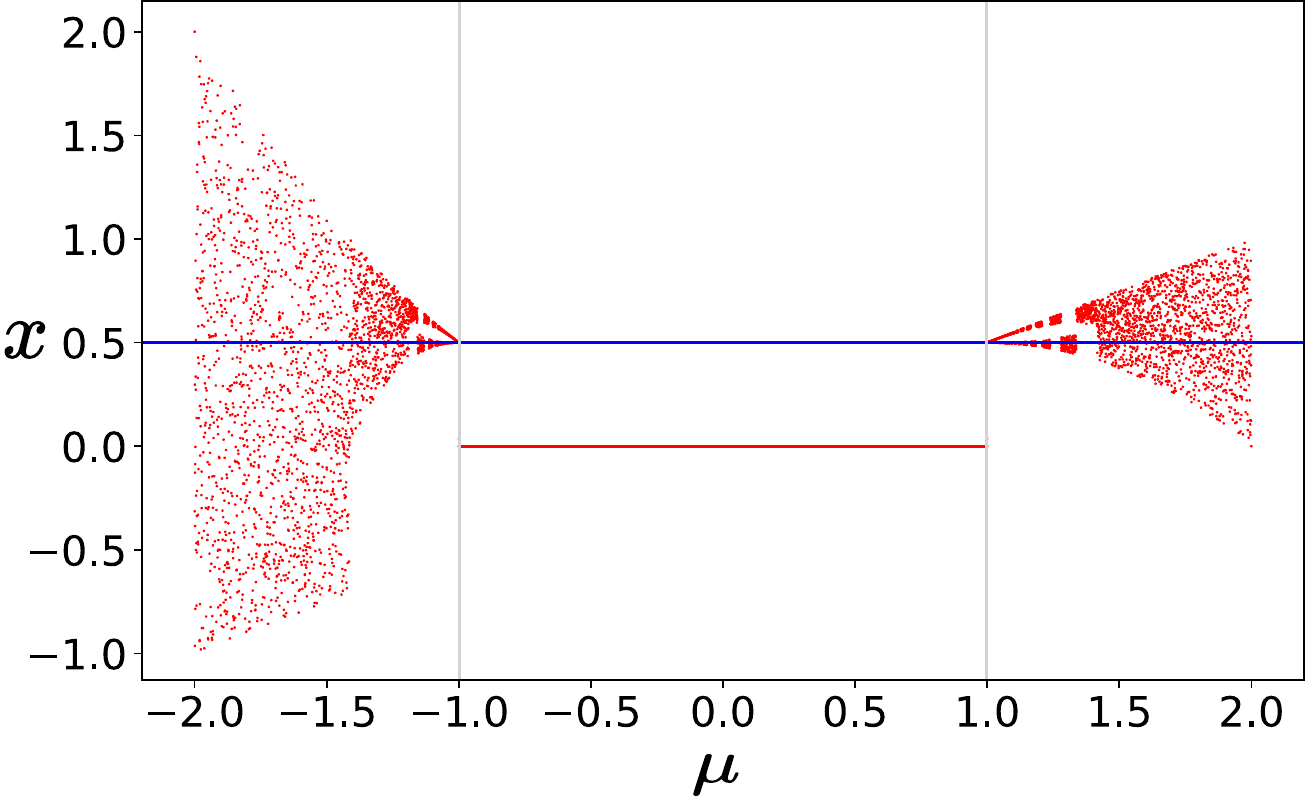}
    \caption{One-parameter border collision bifurcation diagram of \(1D\) tent map showing the parameter \(\mu\) values at which border collision bifurcation occur i.e., \(\mu= -0.9984984984984985\) and \(\mu=0.9984984984984986\) denoted by grey vertical lines.}
    \label{bcbdiagramtentmap}
\end{figure}

Finally, the parameter value \(\mu\) is predicted where the border collision bifurcation occurs, i.e., for which the value of the parameter \(\mu\), \(x\) converges to \(0.5\) or collides the border. And we got two values of \(\mu\) having period \(1\) which shows border collision bifurcation,they are \(\mu=-0.9984984984984985
\) and \(\mu=0.9984984984984986\) as shown in Fig:\ref{bcbdiagramtentmap} in grey vertical lines.

\section{Prediction Using Machine Learning Models} \label{section3}

Machine Learning models are used to predict border collision bifurcation. The aim is to know which model is best for prediction in both Normal Form and Tent Map. Five machine learning models are used. They are:
\begin{enumerate}
    \item Decision Tree Classifier \cite{Priyanka2020}
    \item Logistic Regression \cite{Nick2007}
    \item K-Nearest Neighbors \cite{Kramer2013}
    \item Random Forest \cite{Rigatti2017}
    \item Support Vector Machine \cite{Suthaharan2016}
\end{enumerate}

\subsection{Prediction of the border collision bifurcation of \(1D\) Normal Form Map} \label{sectiondata}

The data is generated in such a way that it simulates equation \eqref{normaleq} across various values of the parameter \(\mu\). It observes how the system's state evolves for each parameter \(\mu\), identifying periodic orbits and recording details such as \(\mu\) value, final state, and period. To create the training and testing set, the original data are divided into features \(X\) and target \(y\). The data is split into \(80\%\) for training and \(20\%\) for testing. All the models are trained to predict the border collision bifurcation from training data. After training, the accuracy of each model is determined by comparing the predictions of the model with the actual results of the experimental data. The models and their corresponding accuracy are shown in table \ref{m&anormal}.

\begin{table}[ht]
\centering
\begin{tabular}{|c|c|c|c|c|c|}
    \hline
    Models & DTC & LR & KNN & RF & SVM \\
    \hline
    Accuracy & 0.965 & 0.685 & 0.955 & 0.975 & 0.685 \\
    \hline
\end{tabular}
\caption{Models: Decision Tree Classifier (DTC), Logistic Regression (LR), K-Nearest Neighbor (KNN), Random Forest (RF), Support Vector Machine (SVM) and their corresponding accuracy.}
\label{m&anormal}
\end{table}

\begin{figure}[!hbtp]
    \centering
    \includegraphics[width=0.8\linewidth]{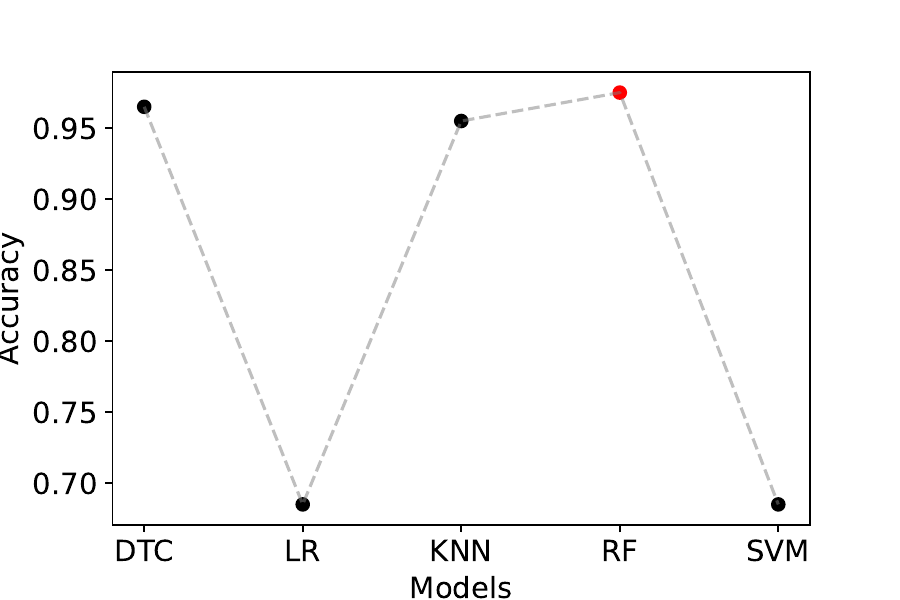}
    \caption{Accuracy vs model plot. The five models: Decision Tree Classifier (DTC), Logistic Regression (LR), K-Nearest Neighbors (KNN), Random Forest (RF) and Support Vector Machine (SVM) and their corresponding accuracy in the prediction of border collision bifurcation.}
    \label{avserrorgraph}
\end{figure}
The figure:\ref{avserrorgraph} is an accuracy vs model plot that compares the accuracy of different machine learning models. The \(x\)-axis represents the models, which are Decision Tree Classifier (DTC), Random Forest (RF) and K-Nearest Neighbors (KNN), and the \(y\)-axis represents the accuracy.
From the figure, we can conclude that Random Forest has high accuracy and is the best machine learning model for predicting the border collision bifurcation of \(1D\) normal form map.
\begin{figure}[!hbtp]
    \centering
    \includegraphics[width=0.8\linewidth]{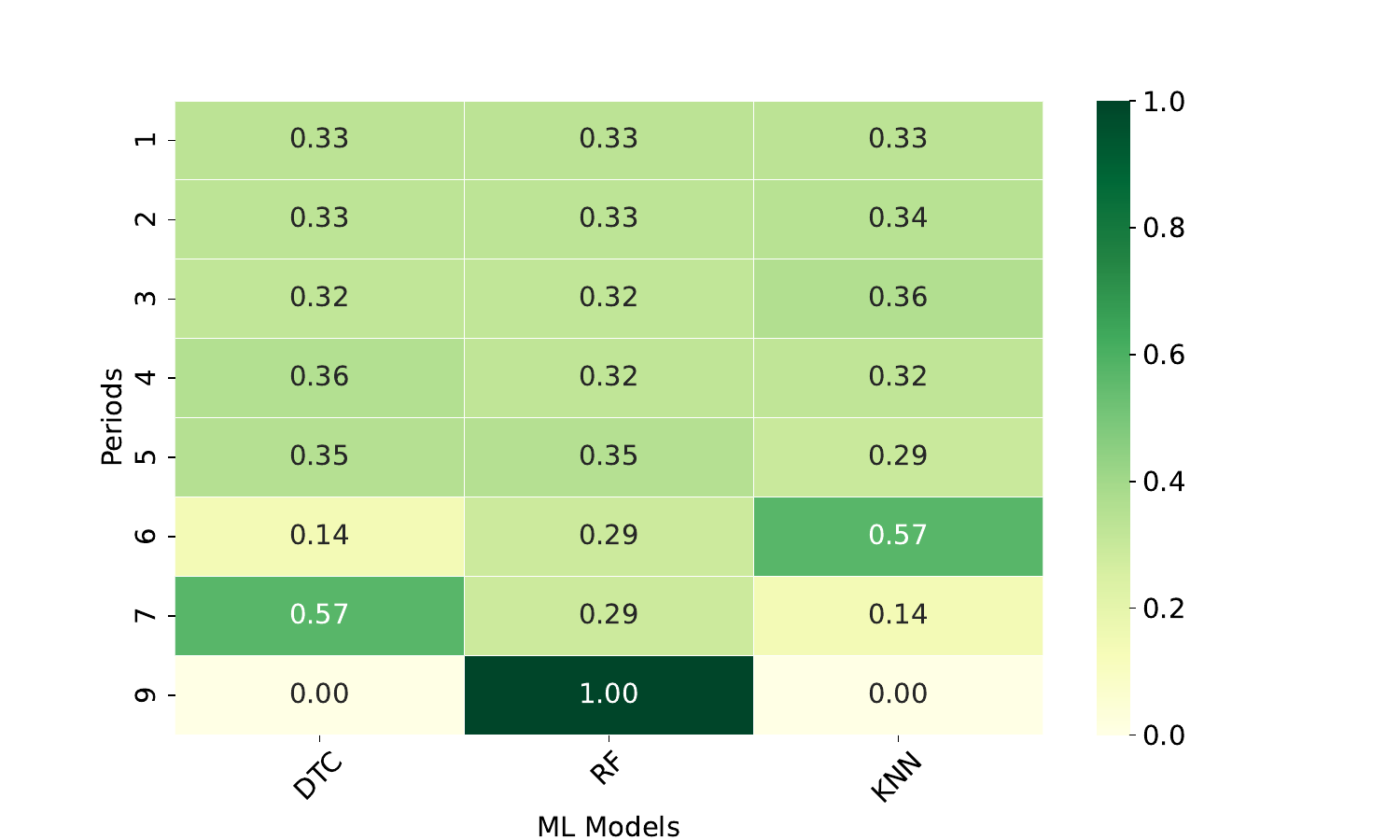}
    \caption{Heatmap of period vs predicted value (\(1D\) Normal Form) from the models(Decision Tree Classifier (DTC), Random Forest (RF) and K-Nearest Neighbors (KNN).}
    \label{heatnormal}
\end{figure}

The heatmap \ref{heatnormal} shows the relative performance of the Decision Tree Classifier (DTC), Random Forest (RF) and K-Nearest Neighbors (KNN) across periodic orbits of different periods. The data is normalised in such a way that all the predicted values will come in one range. Darker colour indicates larger values, and the figure illustrates each model's relative performance or importance over the period. The different colours represent the magnitude of the normalized data values.  In the heatmap, lower values are represented by yellow and higher values by green. The heatmap displays each machine learning model's relative performance or importance (Decision Tree Classifier, Random Forest and K-Nearest Neighbors) across the period because the data has been normalised. Variations in each model's performance are indicated by colour changes.
The light green colours in DTC and RF equivalent cells compared to KNN show that they outperform KNN in periods 1, 2, and 3. Period 4 shows that DTC outperforms KNN, with RF performing similarly. DTC and RF perform comparably in period 5, which is marginally better than KNN. In periods 6 and 7, KNN and RF perform equally, whereas DTC performs poorly. When comparing KNN, RF and DTC respective cells, the darker green colour indicates that RF performs better in period 9.
Thus, we can conclude that the Random Forest (RF) is most reliable and consistent in predicting outcomes across all periods.

\subsection{Prediction of the border collision bifurcation of \(1D\) Tent Map}
The data is generated in such a way that equation \eqref{tenteq} is simulated for the different values of the parameter \(\mu\) and the remaining processes are similar to as mentioned in section \ref{sectiondata}.
The machine learning models and their corresponding accuracy are shown in the table \ref{m&atent}

\begin{table}[ht]
\centering
\begin{tabular}{|c|c|c|c|c|c|}
    \hline
    Models & DTC & LR & KNN & RF & SVM \\
    \hline
    Accuracy & 0.992 & 0.95 & 0.985 & 0.978 & 0.971 \\
    \hline
\end{tabular}
\caption{Models: Decision Tree Classifier(DTC), Logistic Regression(LR), K-Nearest Neighbor(KNN), Random Forest(RF), Support Vector Machine(SVM) and their corresponding accuracy in predicting the border collision bifurcation.}
\label{m&atent}
\end{table}

\begin{figure}[!hbtp]
    \centering
    \includegraphics[width=0.8\linewidth]{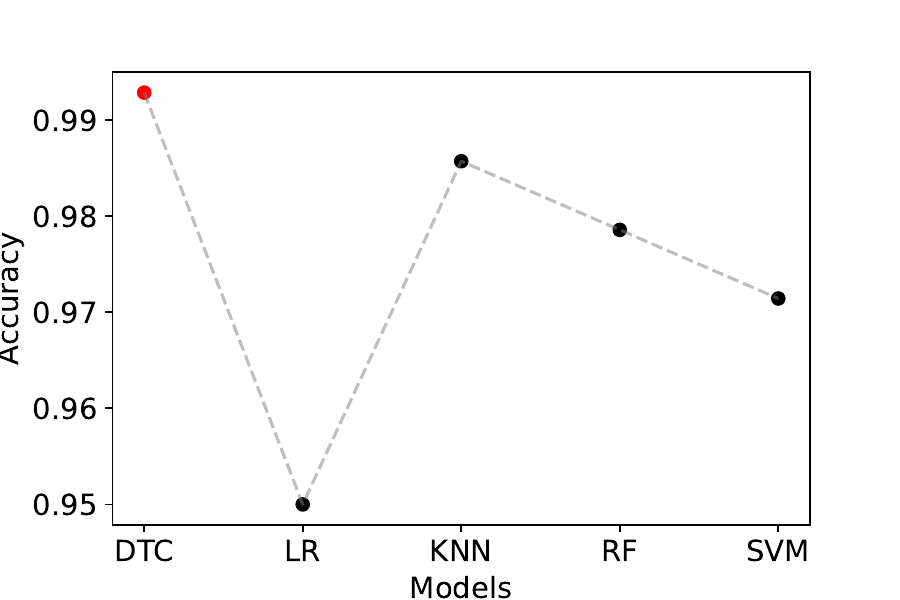}
    \caption{Accuracy vs model plot. The five models: Decision Tree Classifier (DTC), Logistic Regression (LR), K-Nearest Neighbors (KNN), Random Forest (RF) and Support Vector Machine (SVM) and their corresponding accuracy in the prediction of border collision bifurcation.}
    \label{avserrortent}
\end{figure}

The figure:\ref{avserrortent} is an accuracy vs model plot that compares the accuracy of different machine learning models in predicting border collision bifurcation.  The \(x\)-axis represents the models, which are Decision Tree Classifier (DTC), Random Forest (RF) and K-Nearest Neighbors (KNN) and \(y\)-axis represents the accuracy. From the figure, we can conclude that the Decision Tree Classifier have high accuracy and is the best machine learning model for predicting border collision bifurcation of \(1D\) tent map.

Also, from the figure, we can see that apart from the Decision Tree Classifier, Random Forest (RF) and K-Nearest Neighbors (KNN) show accuracy similar to the Decision Tree Classifier. So, we take these three models and plot a period vs predicted value heatmap.

\begin{figure}[!hbtp]
    \centering
    \includegraphics[width=0.8\linewidth]{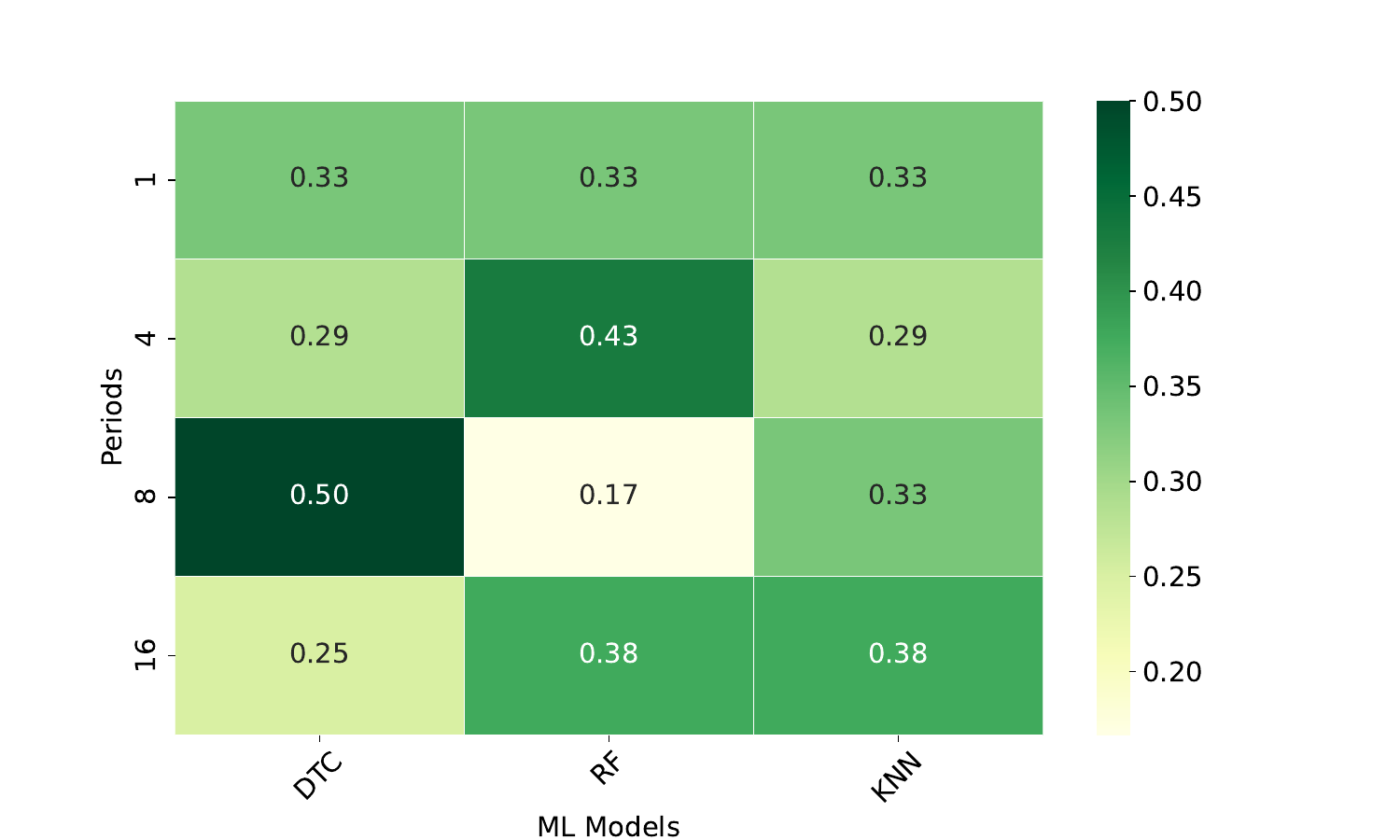}
    \caption{Heatmap of period vs predicted value (\(1D\) tent map) from the models(Decision Tree Classifier (DTC), Random Forest (RF) and K-Nearest Neighbors (KNN).}
    \label{heattent}
\end{figure}

Fig:\ref{heattent} shows the heatmap distribution of three different machine learning models (DTC, RF, and KNN) over five different periods (1, 2, 4, 8, and 16). The data has been normalised so that the sum of each row equals 1, and the colour hues in the heatmap show the relative proportion of each number in a row. All three models have high values, with DTC having the greatest value, as seen in the first period. This implies that while all three models do well in training, DTC has a tiny advantage. It is clear that as we move into higher periods, each model's value distribution varies. Concerning its performance, DTC shows lower values for higher periods. In contrast, the RF and KNN models have lower values for lower periods, but their performance improves with increased periods. The RF model, in particular, performs well during training, as seen by its highest value throughout the most recent period.
Therefore, we can conclude that random forest is best for predicting the periods of the periodic orbit.

Border collision bifurcation can lead to sudden changes like a fixed point after colliding with the boundary, which can bifurcate to a chaotic attractor. So, there is a need to predict the regular and chaotic behaviour of the piecewise smooth map, which will be discussed in the next section.

\section{Classification of Chaotic and Regular Behaviour of Piecewise Smooth Map} \label{section4}
We considered two piecewise smooth maps, \(1D\) Tent Map and \(2D\) Lozi map, for classifying the regular and chaotic dynamical behaviour \cite{Barrio2023}.
\subsection{\(1D\) Tent Map}
The \(1D\) tent map shows both chaotic and regular dynamical behaviour. For classification of this behaviour, the data is generated in such a way that the equation \eqref{tenteq} is simulated for different values of the parameter \(\mu\). The Lyapunov exponent is computed for \( 1000\) randomly selected value of \(r\) in the range of \(-1.5\) to \(1.5\). If the Lyapunov exponent is negative, then the behaviour is regular; if the Lyapunov exponent is positive, then the behaviour is chaotic. Cobweb diagrams are produced for the different values of \(r\). The cobweb diagram is an effective visualization tool for comprehending the behaviour of iterative maps. Each map iteration points \((x, f(x))\) and \((f(x), x)\) are plotted and connected with a line to generate the final result. The cobweb diagrams are generated separately based on the positive (chaotic) or negative (non-chaotic) Lyapunov exponent. The images and their corresponding labels (\(0\) for regular and \(1\) for chaotic) are stored as labels in the dataset.
\begin{figure}[!hbtp]
    \centering
    \includegraphics[width=0.8\linewidth]{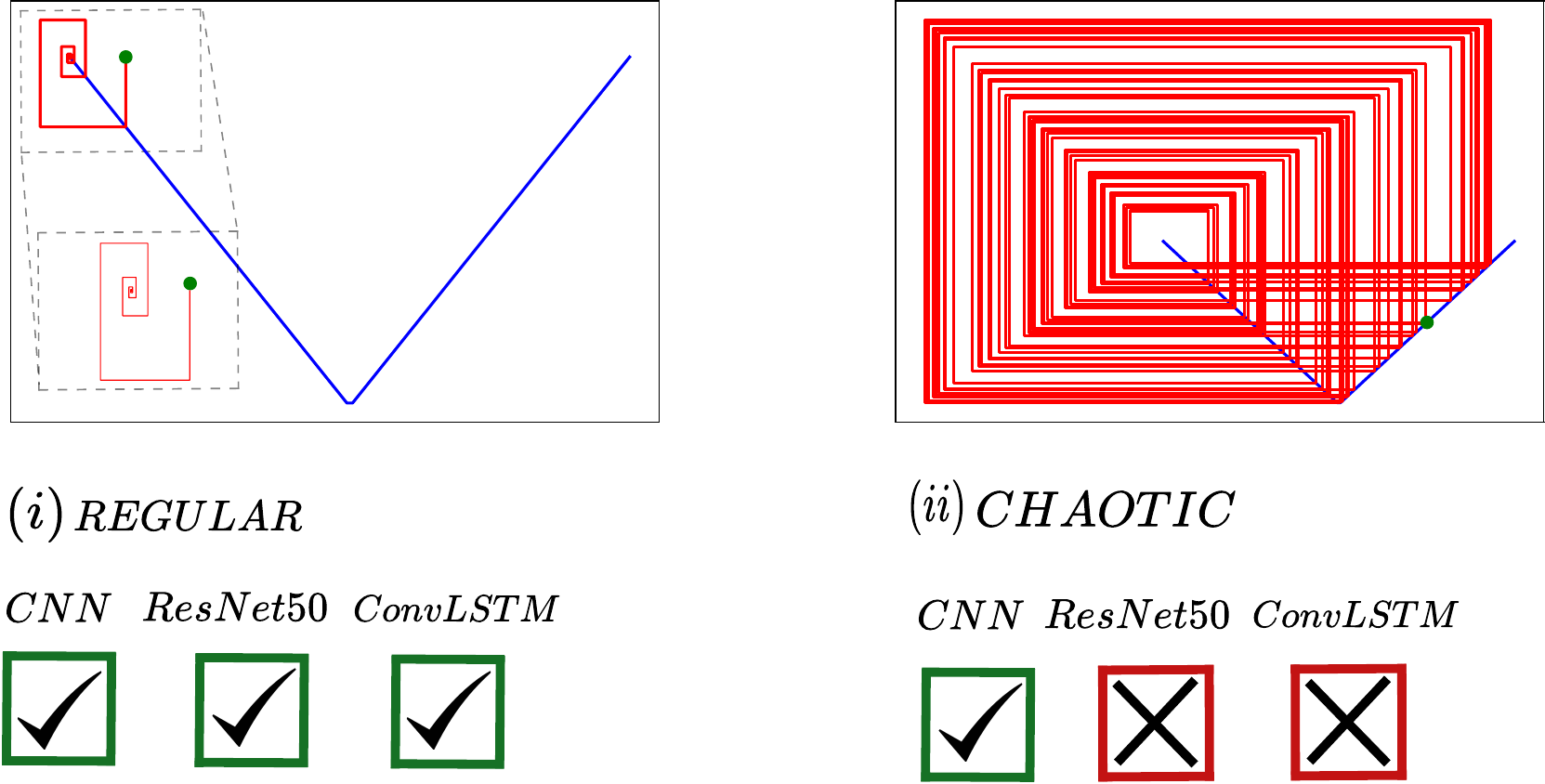}
    \caption{Comparision of regular and chaotic behaviour of Tent Map using cobweb diagram with the three models (CNN, ResNet50 and ConvLSTM). For (i) \(r=-0.53\) (regular) and (ii) \(r=-1.37\) (chaotic).}
    \label{comparison}
\end{figure}
\subsection{\(2D\) Lozi Map}
The \(2D\) Lozi map is a discrete dynamical system defined by the following equation:
\begin{equation}
\begin{aligned}
x_{n+1} &= 1 - a|x_n| + y_n \\
y_{n+1} &= bx_n
\end{aligned}
\label{lozimap}
\end{equation}
 
Here, \((x_n,y_n)\) represents the current state of the system, 
\((x_{n+1},y_{n+1})\) denotes the next state, and \(a\) and \(b\) are parameters that determine the behavior of the system.

The data is generated by simulating the system equation \eqref{lozimap} for \(a\in(-0.1, 1.7)\), while keeping the parameter \(b=0.5\) constant. For each value of \(a\), the system iterates the map function for \(10000\) iterations, starting with random initial values for \(x\) and \(y\) trajectories are shown. To measure the system's sensitivity to initial conditions, the Lyapunov exponents are calculated for the longer duration of each iteration.

The chaotic (positive exponent) or regular (negative exponent) dynamical behaviour of the \(2D\) Lozi map is determined using the Lyapunov exponent. Additionally, the trajectory is plotted for each \(a\) to create a phase portrait, visually representing the system's state evolution. The Lyapunov exponent categorises the phase portraits as chaotic or regular. A dataset of trajectories and Lyapunov exponents is produced by repeating this approach over the entire range of \(a\).

\begin{figure}[!hbtp]
    \centering
    \includegraphics[width=0.8\linewidth]{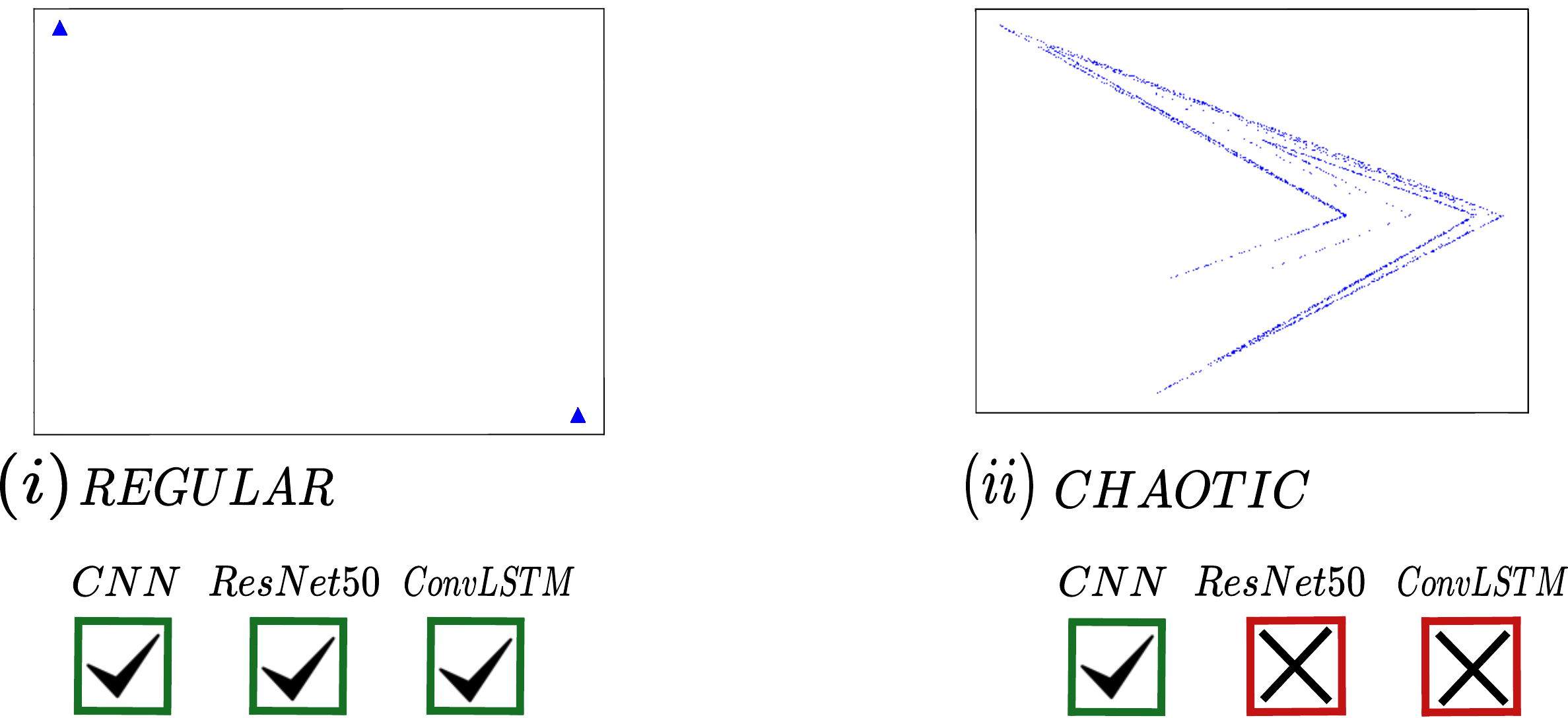}
    \caption{Comparision of regular and chaotic behaviour of Lozi Map using phase portrait with the three models (CNN, ResNet50 and ConvLSTM). For (i) \(r=1.10\) (regular) and (ii) \(r=1.68\) (chaotic).}
    \label{comparisonlozi}
\end{figure}

\subsection{Deep Learning Models}

The dataset is splitted into training \((70\%)\) and testing \((30\%)\).
Three deep learning architectures are used to classify chaotic and regular dynamical behaviour. They are Convolutional Neural Network (CNN), ResNet50 and ConvLSTM.

\begin{itemize}
    \item Convolutional Neural Network (CNN)

    The first layer is a convolutional layer with \(32\) neurons, \(3 \times 3\) kernel, and the ReLu activation function is used. The second layer is a max pooling layer of \(2 \times 2\), which reduces the spatial dimensions of the input neuron. The third layer is a flattening layer, which reshapes the \(2\)D output into a \(1\)D array. The fourth layer is a fully connected layer with \(256\) neurons and with ReLU activation function. The fifth layer has a dropout rate of \(0.5\), which prevents overfitting. The sixth layer is a fully connected layer with \(512\) neurons and the ReLU activation function used. The final layer is a fully connected layer with \(2\) neurons and a softmax activation function, which produces the output predictions for the model and converts the output values into possibilities that sum to \(1\), allowing the model to predict multiple classes.
\begin{figure}[!hbtp]
    \centering
    \includegraphics[width=0.8\linewidth]{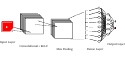}
    \caption{CNN architecture showing how regular and chaotic behaviour are classified. Cobweb diagram \(r=-1.37\) is given as input and prediction (chaotic \((1)\) or regular \((0)\) ) are given as output. }
    \label{cnncobweb}
\end{figure}

    \item ResNet50

    Defines a sequential model that includes multiple layers. The first layer is a pre-trained ResNet50 model used as a feature extractor and set to be non-trainable. The second layer is a global common pooling layer, which reduces the spatial dimensions of the input. The third and fourth layers are fully connected layers with \(256\) neurons and a ReLU activation function. The fifth layer has a dropout rate of \(0.5\), which prevents overfitting. The final layer is a fully connected layer with \(2\) neuron and a sigmoid activation function, producing the binary type classification output predictions. 

    \item ConvLSTM 

    Defines a sequential model that includes multiple layers. The first layer is a TimeDistributed convolutional layer with \(32\) neurons and a kernel size of \((3 \times 3)\), which applies the convolution operation to the step of the input collection. The second layer is a TimeDistributed max pooling layer with a pool length of \((2 \times 2)\), which reduces the spatial dimensions of the input at every time step. The third layer is a TimeDistributed flattening layer, which reshapes the \(3\)D tensor into a \(2\)D tensor. The fourth layer is an LSTM layer with \(64\) units, which learns patterns over time in the input sequence. The final layer is a fully connected layer with \(2\) neuron and a sigmoid activation function, which produces the output predictions for the binary classification task.
\end{itemize}
 The Adam optimiser and the binary cross-entropy loss function are used for the construction of all models. A batch size of \(32\), the model is trained for \(50\) epochs. After training, the accuracy is evaluated.
 
Table \ref{tabletent} and \ref{tablelozi1} shows the accuracy and loss of test, validation, and training datasets for the three deep learning methods (CNN, ResNet50, and ConvLSTM). When we compare the test data results, CNN appears to provide better results than the other two networks. In addition, we calculated the accuracy in the test set of regular and chaotic samples, yielding the results shown in Table \ref{tableltent2} and \ref{tablelozi2}. The fact that both regular and chaotic sample percentages are \(100\%\) indicates that CNN has accurately learned the key characteristics of both kinds of dynamical behaviour.

\begin{table}
    \centering
    \begin{tabular}{cccccccc}
    \toprule
    \toprule
    \multicolumn{1}{c}{} & \multicolumn{2}{c}{\textbf{CNN}} & \multicolumn{2}{c}{\textbf{ResNet50}} & \multicolumn{2}{c}{\textbf{ConvLSTM}}\\
    \cmidrule(rl){2-3} \cmidrule(rl){4-5} \cmidrule(rl){6-7}
    \textbf {} & {Loss} & {Accuracy } & {Loss} & {Accuracy  } & {Loss} & {Accuracy }\\
    \midrule
    Training & 0.0001 & 1.0 & 0.1741 & 1.0 & 0.6323 & 0.6739 \\
    Validation & 0.000002 & 1.0 & 0.0977 & 0.9955 &  0.6739 & 0.6797 \\
    Test & 0.00000045 & 1.0 &  0.0866 &  0.9960 &  0.6215 & 0.6901 \\
    \bottomrule
    \bottomrule
    \end{tabular}
    \caption{Loss and Accuracy (\%) values for different deep learning models, CNN, ResNet50 and ConvLSTM of \(1D\) Tent Map. }
    \label{tabletent}
\end{table}

\begin{table}
    \centering
    \begin{tabular}{ccccccccc}
    \toprule
    \toprule
    \multicolumn{1}{c}{} & \multicolumn{2}{c}{\textbf{CNN}} & \multicolumn{2}{c}{\textbf{ResNet50}} & \multicolumn{2}{c}{\textbf{ConvLSTM}}\\
    \cmidrule(rl){2-3} \cmidrule(rl){4-5} \cmidrule(rl){6-7}
    \textbf {} & {Loss} & {Accuracy } & {Loss} & {Accuracy  } & {Loss} & {Accuracy }\\
    \midrule
    Training & 0.0222 & 1.0 & 0.5607 & 0.7216 & 0.6475 & 0.6853 \\
    Validation & 0.0114 & 1.0 & 0.2903 & 0.1 &  0.6327 & 0.7179 \\
    Test & 0.0113 & 1.0 &  0.6367 &  0.6851 &  0.6609 & 0.6363 \\
    \bottomrule
    \bottomrule
    \end{tabular}
    \caption{Loss and Accuracy (\%) values for different deep learning models, CNN, ResNet50 and ConvLSTM of \(2D\) Lozi map.}
    \label{tablelozi1}
\end{table}
\begin{table}
 \centering
    \begin{tabular}{ccccccccc}
    \toprule
    \toprule
    \multicolumn{1}{c}{} & \multicolumn{1}{c}{\textbf{CNN}} & \multicolumn{1}{c}{\textbf{ResNet50}} & \multicolumn{1}{c}{\textbf{ConvLSTM}}\\
    \cmidrule(rl){2-2} \cmidrule(rl){3-3} \cmidrule(rl){4-4}
    \textbf {}  & {Accuracy } & {Accuracy  }  & {Accuracy  }\\
    \midrule
    Chaotic & 1.0 & 0.987 & 0 \\
    Regular & 1.0 & 1.0 & 1.0  \\
    \bottomrule
    \bottomrule
    \end{tabular}
    \caption{Accuracy of different deep learning models, CNN, ResNet50 and ConvLSTM for chaotic and regular behaviours of Tent Map.}
    \label{tableltent2}
\end{table}
    
\begin{table}
 \centering
\centering
    \begin{tabular}{ccccccccc}
    \toprule
    \toprule
    \multicolumn{1}{c}{} & \multicolumn{1}{c}{\textbf{CNN}} & \multicolumn{1}{c}{\textbf{ResNet50}} & \multicolumn{1}{c}{\textbf{ConvLSTM}}\\
    \cmidrule(rl){2-2} \cmidrule(rl){3-3} \cmidrule(rl){4-4}
    \textbf {}  & {Accuracy } & {Accuracy  }  & {Accuracy  }\\
    \midrule
    Chaotic & 1.0 & 0 & 0 \\
    Regular & 1.0 & 1.0 & 1.0  \\
    \bottomrule
    \bottomrule
    \end{tabular}
    \caption{Accuracy of different deep learning models, CNN, ResNet50 and ConvLSTM for chaotic and regular behaviours of Lozi Map.}
    \label{tablelozi2}
\end{table}

\begin{figure}[!hbtp]
    \centering
    \includegraphics[width=1\linewidth]{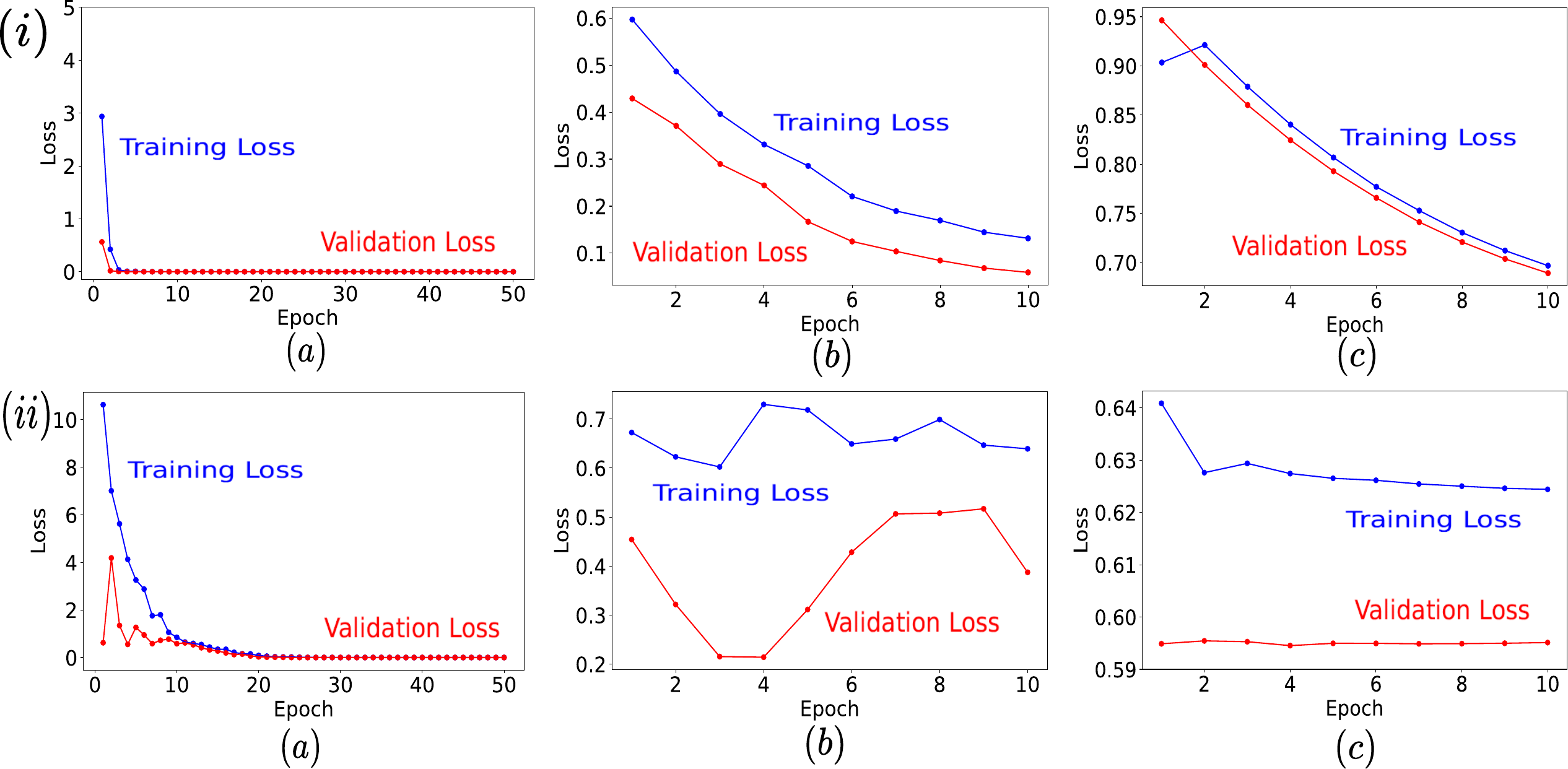}
    \caption{Epoch vs Loss in training and validation datasets of \((i)\) Tent Map and \((ii)\) Lozi Map for \((a)\) CNN, \((b)\) ResNet50, \((c)\) ConvLSTM; There are two lines, blue representing the training loss and red representing the validation loss.}
    \label{evslrvsc}
\end{figure}

\begin{figure}[!hbtp]
    \centering
    \includegraphics[width=1\linewidth]{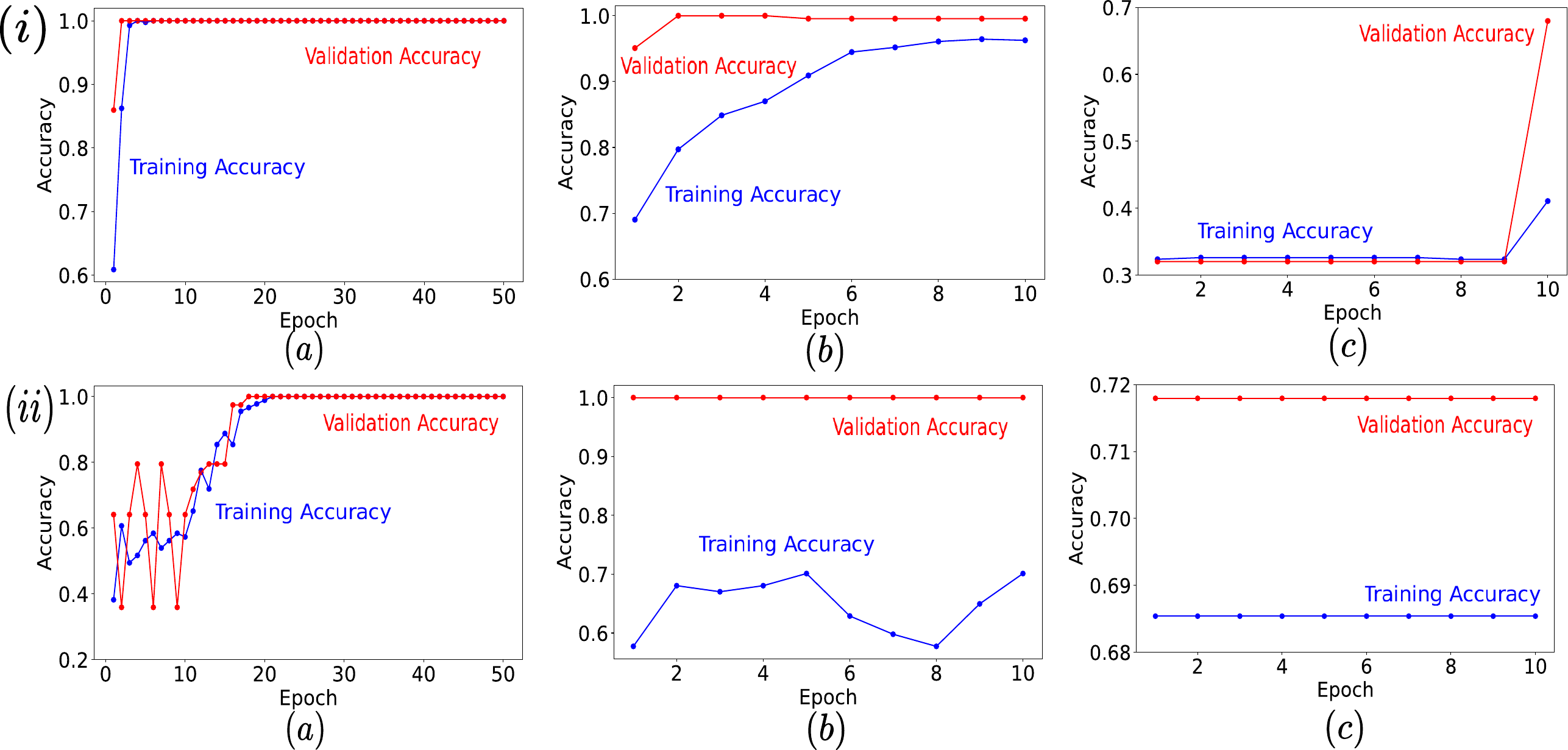}
    \caption{Epoch vs Accuracy in training and validation datasets of \(i\) Tent Map and \(ii\) Lozi Map for \((a)\) CNN, \((b)\) ResNet50, \((c)\) ConvLSTM; There are two lines, blue representing the training accuracy and red representing the validation accuracy.}
    \label{evsarvsc}
\end{figure}

We discussed the regular and chaotic behaviour of \(1D\) and \(2D\) piecewise smooth maps. Now, we will discuss the hyperchaotic behaviour of the \(3D\) piecewise smooth map.

\section{Classification of Chaotic and Hyperchaotic Behaviour of Piecewise Smooth Map} \label{section5}

\subsection{\(3D\) Piecewise Smooth Map}
The three-dimensional piecewise smooth normal form map \cite{Patra2020} is given by
\begin{equation}
X_{n+1}=
    \begin{cases}
        A_lX_n+\mu C  ,& \text{if } X_n \le 0\\
        A_rX_n+\mu C , & \text{if } X_n \ge 0
    \end{cases}
\end{equation}

where \(X_n=(x_n,y_n,z_n)^T \in \mathbb{R}_3\), \(C=(1,0,0)^T \in \mathbb{R}_3 \), \(A_l\) and \(A_r\) are real valued \(3\times3\) matrices
\(A_l=\)
$\begin{pmatrix}
\tau_l & 1 & 0\\
-\sigma_l & 0 & 1\\
\delta_l & 0 & 0
\end{pmatrix}$
and \(A_r=\)
$\begin{pmatrix}
\tau_r & 1 & 0\\
-\sigma_r & 0 & 1\\
\delta_r & 0 & 0
\end{pmatrix}$

\begin{figure}[!hbtp]
    \centering
    \includegraphics[width=0.8\linewidth]{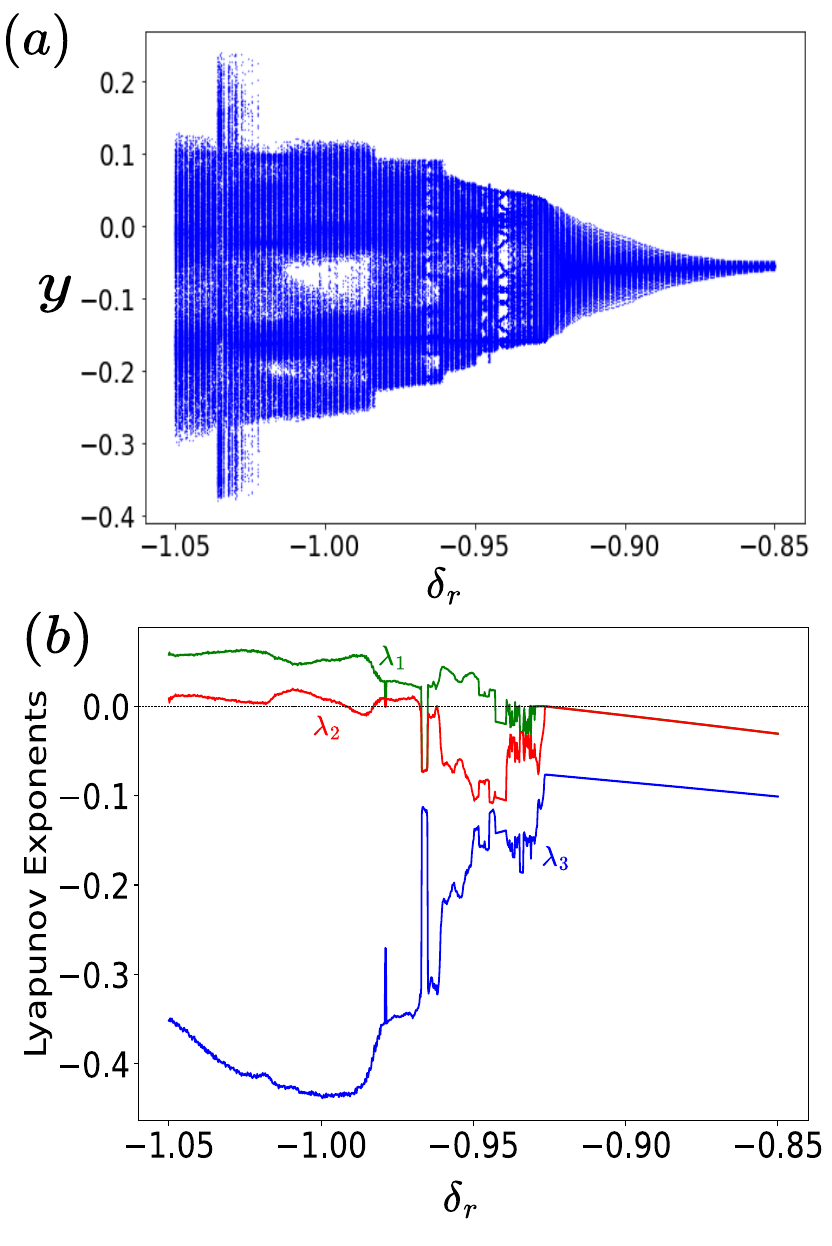}
    \caption{(a)Bifurcation diagram and (b)Lyapunov exponent spectrum, where green shows the \(\lambda_1\) values, red shows the \(\lambda_2\) values, and blue shows the \(\lambda_3\) values and \(\delta_r\in(-1.05,-0.85)\).}
    \label{3dLE}
\end{figure}

For a range of \(\delta_r\) values \((-1.05\) to \(-0.85)\),  the equation is simulated to produce data. For each \(\delta_r\), there are defined matrices \(A_l\) and \(A_r\) that depend on parameters \(\tau_l=-0.5\), \(\sigma_l=0.95\), \(\delta_l=0.2\), \(\tau_r=0.8\), \(\sigma_r=-0.6\) and \(\mu=0.1\), which determine the dynamics of the system. After being initialised at \(x\), it iterates through \(10000\) steps where the state is updated by the system's map function, calculating Lyapunov exponents. It labels the system’s behaviour according to the exponents. If all \(\lambda_1\), \(\lambda_2\) and \(\lambda_3\) are less than 0, then the label is 0 (regular), if any one of \(\lambda_1\), \(\lambda_2\) and \(\lambda_3\) is greater than 0, then the label is 1 (chaotic) and if any two of \(\lambda_1\), \(\lambda_2\) and \(\lambda_3\) are positive, then the label is 2 (hyperchaos) as shown in the figure \ref{3dLE}.

The Deep Learning models used for the classification:
\begin{itemize}
    \item FeedForward Neural Network

    The model’s architecture has dense layers containing the ReLU activation function and dropout layers to decrease overfitting. The model begins with an input layer of \(64\) neurons and ReLU activation; each subsequent pair of dense layers maintains this pattern, progressively reducing neuron count to \(32, 16,\) and finally outputting \(4\) neurons with a softmax activation function.
    
    \item Long short-term memory (LSTM)

    The model´s architecture is defined by a set of layers that consist of different numbers of neurons to capture sequential patterns efficiently. It begins with an LSTM layer with an input of \(138\) neurons. After the first layer, there are dropout layers after each LSTM layer, which have been set at \(20\%\) to avoid overfitting. The following dense layers have \(64\) and \(32\) neurons, respectively, using ReLU activation functions to allow a non-linear nature and to extract higher-level features from LSTM outputs. The last dense layer includes 4 neurons with a softmax activation function.

    \item Recurrent Neural Network

    The model begins with several numbers of SimpleRNN layers, each having \(64\) neurons used for detecting temporal patterns. Dropout layers with a dropout rate of \(20\%\) come after each RNN layer in order to avoid overfitting. When all RNN layers are passed, a dense layer is added to this model. It contains \(32\) neurons and comes with a ReLU activation function. The final layer has \(4\) neurons which use the softmax activation function.
    
\end{itemize}
All models are trained using an optimizer called the Adam optimiser and sparse categorical cross-entropy loss. Over \(100\) epochs with a batch size of \(32\) are used during the training. Once training is complete, the accuracy of the model is evaluated.
\begin{figure}[!hbtp]
    \centering
    \includegraphics[width=0.8\linewidth]{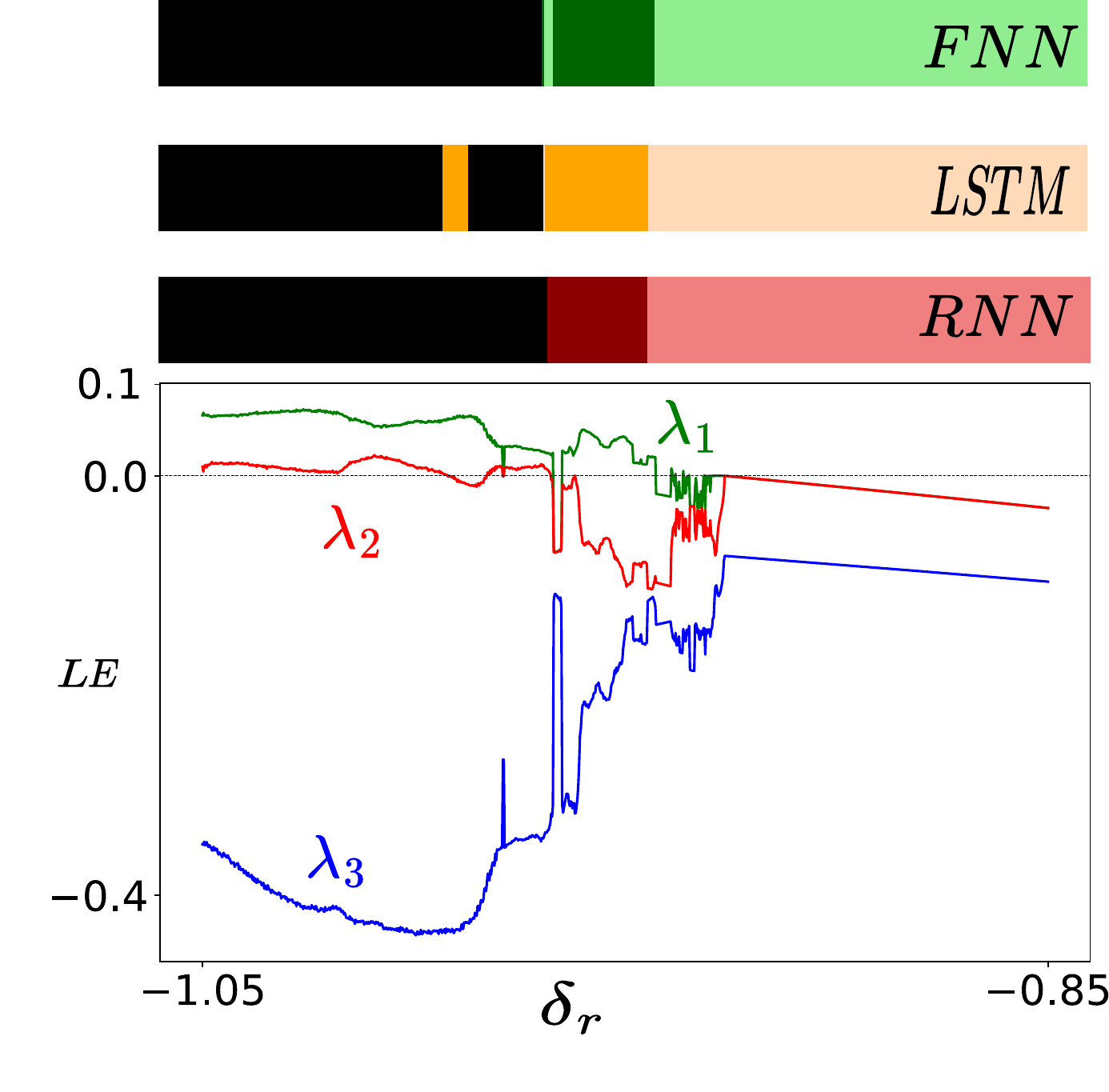}
    \caption{Comparison of predicted behaviour (FNN in green, LSTM in orange and RNN in red) of \(3D\) piecewise smooth map with Lyapunov spectrum.}
    \label{barcomp}
\end{figure}

\begin{figure}[!hbtp]
    \centering
    \includegraphics[width=0.8\linewidth]{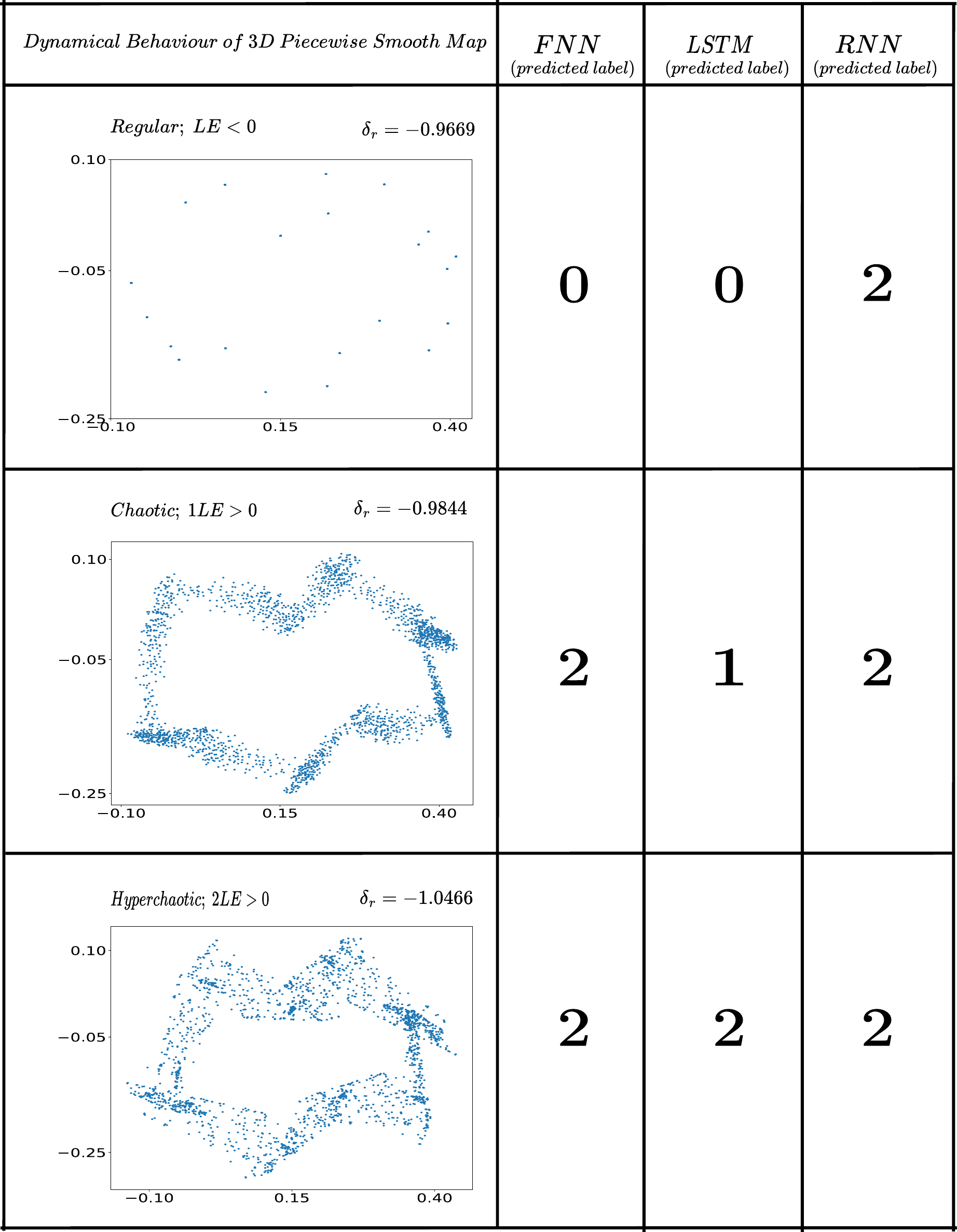}
    \caption{Comparison of predicted labels (regular \((0)\), chaotic \((1)\), hyperchaotic \((2)\)) of three models (FNN, LSTM, RNN) with the help of phase portraits.  }
    \label{ppcomp}
\end{figure}

The table \ref{3DSMOOTHTABLE} shows that the LSTM model performs better when compared to the Feed Forward Neural Network (FNN) and the Recurrent Neural Network (RNN). It has the highest training accuracy with \(92.00\%\), validation with \(93.50\%\) as well, as testing accuracy of \(93.50\%\) respectively; besides, it experiences the lowest losses, which are coincidentally equal to \(0.2067\) for both training set error rates and \(0.2048\) for each other set. This can be interpreted as a sign that the LSTM model is better able to generalise beyond the seen data points while at the same time producing a more precise hypothesis.

\begin{table}
    \centering
    \begin{tabular}{ccccccccc}
    \toprule
    \toprule
    \multicolumn{1}{c}{} & \multicolumn{2}{c}{\textbf{FNN}} & \multicolumn{2}{c}{\textbf{LSTM}} & \multicolumn{2}{c}{\textbf{RNN}}\\
    \cmidrule(rl){2-3} \cmidrule(rl){4-5} \cmidrule(rl){6-7}
    \textbf {} & {Loss} & {Accuracy } & {Loss} & {Accuracy  } & {Loss} & {Accuracy }\\
    \midrule
    Training & 0.2235 & 0.9187 & 0.2067 & 0.9200 & 0.2305 & 0.9137\\
    Validation & 0.2406 & 0.8899 & 0.2048 & 0.9350 &  0.2327 & 0.8899 \\
    Test & 0.2406 & 0.8899 &  0.2048 &  0.9350 &  0.2327 & 0.8899 \\
    \bottomrule
    \bottomrule
    \end{tabular}
    \caption{Loss and Accuracy (\%) values of training, validation, and test dataset of \(3D\) Piecewise Smooth Map for different deep learning models.}
    \label{3DSMOOTHTABLE}
\end{table}

\begin{figure}[!hbtp]
    \centering
    \includegraphics[width=1\linewidth]{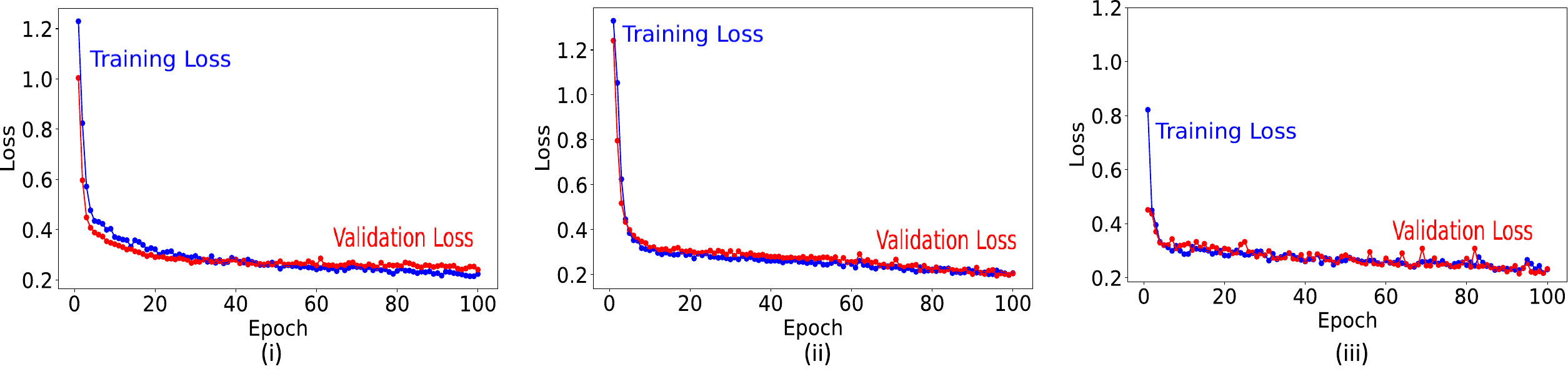}
    \caption{Epoch vs Loss in training and validation datasets of \(3D\) Piecewise Smooth Map for \((i)\) FNN, \((ii)\) LSTM, \((iii)\) RNN; There are two lines, blue representing the training loss and red representing the validation loss.}
    \label{evslcvsh}
\end{figure}

\begin{figure}[!hbtp]
    \centering
    \includegraphics[width=1\linewidth]{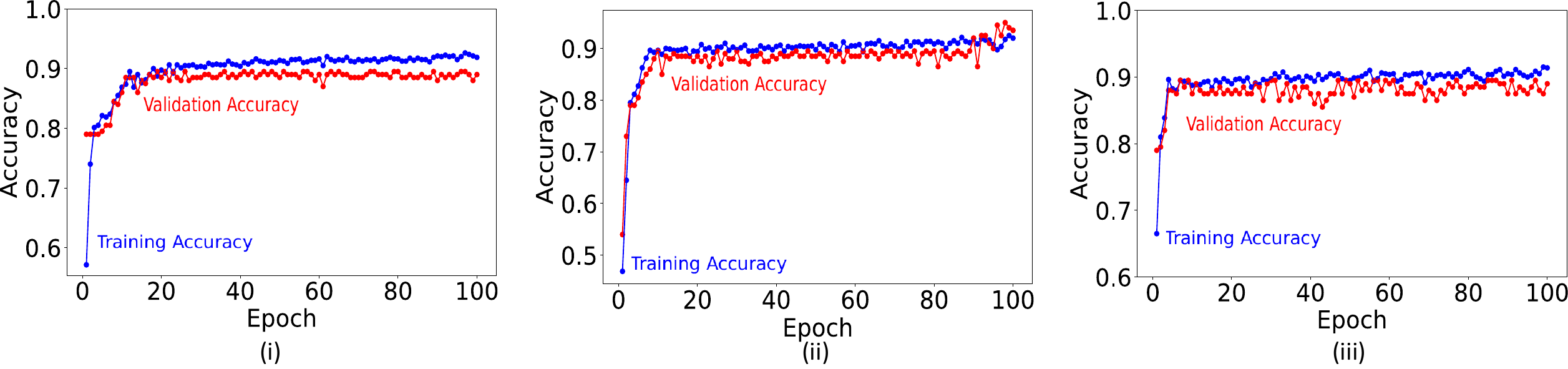}
    \caption{Epoch vs Accuracy in training and validation datasets of \(3D\) Piecewise Smooth Map for \((i)\) FNN, \((ii)\) LSTM, \((iii)\) RNN; There are two lines, blue representing the training loss and red representing the validation loss.}
    \label{evsacvsh}
\end{figure}

\section{Two Parameter Charts} \label{section6}
\begin{figure}[!hbtp]
    \centering
    \includegraphics[width=1\linewidth]{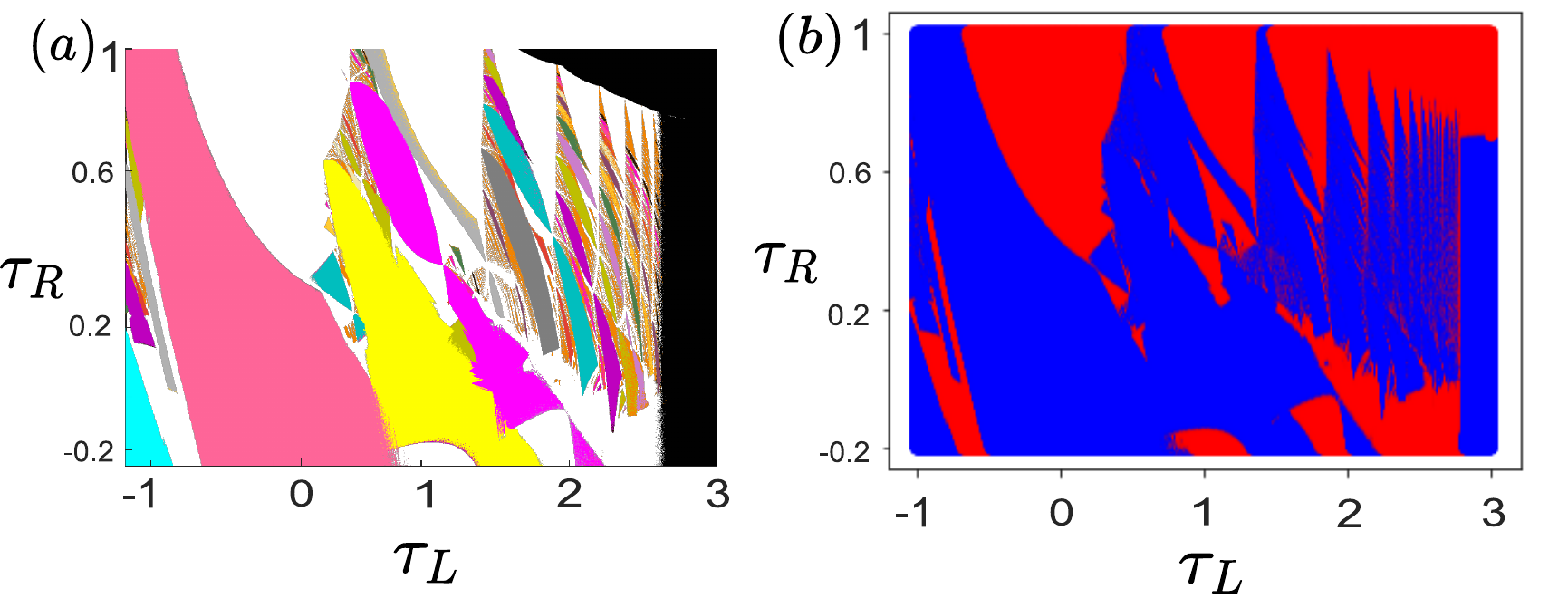}
    \caption{Biparametric plane \((\tau_L,\tau_R)\) of \(2D\) normal form map where \(\tau_l\in(-1,3)\) and \(\tau_R\in(-0.2,1)\) (a)Image showing periods in different colours (b)Blue (regular) and red (chaotic).}
    \label{2para}
\end{figure}

Two parameter charts are used in the field of nonlinear dynamics and chaos theory to study the stability and behaviour of the dynamical systems \cite{McGuinness2004}. They are used to visualize how the system's behaviour changes when the parameter varies.
Here, we used a \(2D\) border collision bifurcation normal form map for studying the behaviour using a two-parameter chart.
The border collision bifurcation normal form map is defined as follows:

\begin{equation}
[x, y] \mapsto
\begin{cases}
\begin{bmatrix}
\tau_L x + y + \mu \\
-\delta_L x
\end{bmatrix}, & x \leq 0, \\[10pt]
\begin{bmatrix}
\tau_R x + y + \mu \\
-\delta_R x
\end{bmatrix}, & x > 0
\end{cases}
\label{eq:piecewise}
\end{equation}

where \(\delta_L=2\), \(\delta_R=-0.2\) and \(\mu=-1\).
The data is generated by simulating the equation \ref{eq:piecewise}, where \(\tau_L\in(-1,3)\) and \(\tau_R\in(-0.2,1)\) for 1000 points. The Lyapunov exponents are calculated and labelled as regular(0) and chaotic(1). A biparametric plane is plotted using these labels as shown in the figure \ref{2para}.
Figure: \ref{2para} shows the biparametric plane \((\tau_L,\tau_R)\) of \(2D\) normal form map where \(\tau_l\in(-1,3)\) and \(\tau_R\in(-0.2,1)\).
Figure: \ref{2para}(a) is the original image where coloured regions are periodic and white regions are chaotic. Then we converted figure \ref{2para}(a) to figure: \ref{2para}(b) biparametric plane where regular regions are shown in blue and chaotic regions in red.
For the training, \(\tau_L\in(0,2)\) and \(\tau_R\in(0.6,1)\) are taken. As generating the two parametric charts computationally is time-consuming, we used Recurrent Neural Network (RNN) and Long Short-Term Memory (LSTM) to predict the labels.
The data is splitted into training \(80\%\) and testing(20\%) datasets.
\begin{itemize}
    \item Recurrent Neural Network (RNN)

    The model begins with a SimpleRNN layer with 256 neurons. After the first layer, there are more SimpleRNN layers, gradually decreasing the number of neurons from 256 to 2. The models then include flatten layers and dense layers with 256 neurons and the ReLU activation function. Another dense layer with 64 neurons is added before the final output layer, which uses the softmax activation function.

    \item Long Short-Term Memory (LSTM)

     The model begins with the LSTM layer having 256 neurons. This layer is followed by several more LSTM layers, each gradually reducing the number of neurons from 256 to 2. After the series of LSTM layers, the models include a flatten layer. The dense layers are added to perform the final classification, which includes two layers with 256 and 128 neurons, respectively. Another dense layer with 64 neurons and the final output layer uses the softmax activation function for the classification task.
\end{itemize}
Both models are compiled with the Adam optimiser and categorical cross-entropy loss and trained for 100 epochs with a batch size of 64. After training, the accuracy of the models is calculated.

\begin{figure}[!hbtp]
    \centering
    \includegraphics[width=1\linewidth]{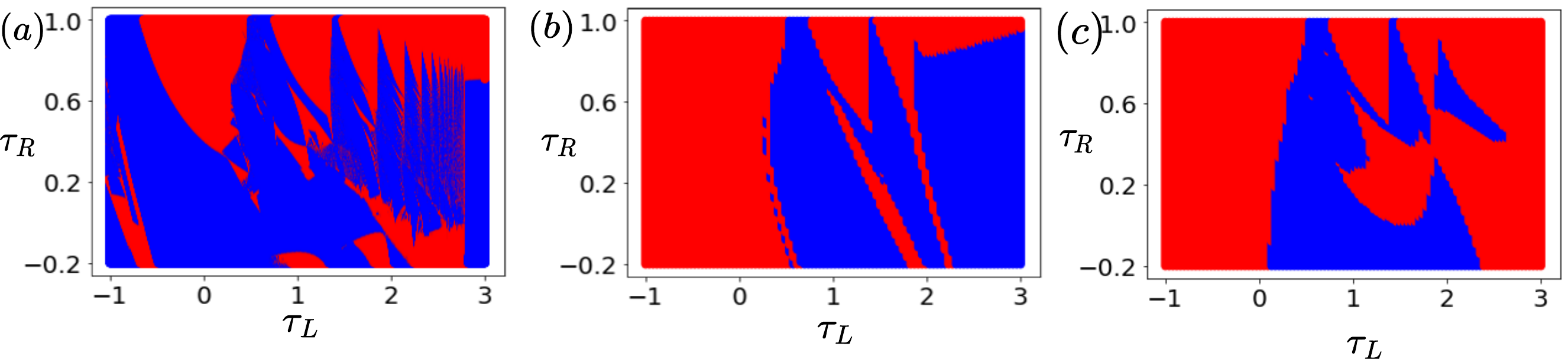}
    \caption{Biparametric plane \((\tau_L,\tau_R)\) of \(2D\) normal form map where \(\tau_l\in(-1,3)\) and \(\tau_R\in(-0.2,1)\) (a)image plotted using actual values, image plotted using predicted values (b) (RNN model) (c) LSTM model}
    \label{2para100}
\end{figure}

The labels are predicted for the range of parameter \(\tau_L\in(-1,3)\) and \(\tau_R\in(-0.2,1)\) and plotted in such a way that regular region in blue and chaotic region in red as shown in the figure: \ref{2para100}(b) using RNN and figure: \ref{2para100}(c) using LSTM.
The accuracy of RNN and LSTM are 0.9647 and 0.9817, respectively, which shows that LSTM is more accurate than RNN. 

\section{Conclusion} \label{section7}
In this paper, we have used deep learning models for the prediction and classification of the dynamics of the piecewise smooth maps like  \(1D\) normal form map, \(1D\) tent map, \(2D\) Lozi Map, \(3D\) piecewise smooth map. The machine learning models used include Decision Tree Classifier, Logistic Regression, K-Nearest Neighbor, Random Forest and Support Vector Machine for predicting border collision bifurcation of \(1D\) normal form map and \(1D\) tent map. Moreover, deep learning models such as Convolutional Neural Network (CNN), ResNet50 and ConvLSTM are used for classifying regular and chaotic dynamical behaviour of \(1D\) tent map and \(2D\) lozi map. Feedforward Neural Network (FNN), Long Short-Term Memory(LSTM) and Recurrent Neural Network(RNN) are used for classifying chaotic and hyperchaotic dynamical behaviour in \(3D\) piecewise smooth map. Finally, we reconstructed the two parametric charts of \(2D\) border collision bifurcation normal form map using deep learning models like Recurrent Neural Network (RNN) and Long Short-Term Memory (LSTM).

We have used the Lyapunov exponents, cobweb diagrams, and phase portraits to classify regular, chaotic, and hyperchaotic dynamical behaviour during data generation. For the prediction of border collision bifurcation of \(1D\) normal form map, random forest shows more accuracy, and in \(1D\) tent map, the decision tree classifier shows more accuracy. The convolutional neural network (CNN) is suitable for classifying regular and chaotic dynamical behaviour. LSTM is the most accurate among other deep learning models for classifying chaotic and hyperchaotic behaviour. Also, LSTM is more precise than RNN in reconstructing the two parametric charts.
Future scopes include applying deep learning to predict the cobweb symbolisses, as doing it manually is very hard. Deep learning models can also predict the border collision bifurcation of higher periods.

\section*{Acknowledgements}
We acknowledge insightful discussions with Thomas M. Bury regarding deep learning and prediction of codimension-one bifurcations in smooth maps.

\section*{Conflict of interest}
 The authors declare that they have no conflict of interest.

 \section*{Data Availability Statement}
 The data that support the findings of this study are available within the article.

\bibliographystyle{unsrt}
\bibliography{Arxiv}

\end{document}